\documentclass[10pt,twocolumn,letterpaper]{article}

\usepackage[pagenumbers]{cvpr} 

\usepackage{graphicx}
\usepackage{amsmath}
\usepackage{amssymb}
\usepackage{booktabs}
\usepackage{url}
\usepackage{algorithm, algorithmic}
\usepackage{caption}
\usepackage{subcaption}
\usepackage{multirow}
\usepackage{xcolor}
\usepackage[pagebackref,breaklinks,colorlinks]{hyperref}

\usepackage[capitalize]{cleveref}
\crefname{section}{Sec.}{Secs.}
\Crefname{section}{Section}{Sections}
\Crefname{table}{Table}{Tables}
\crefname{table}{Tab.}{Tabs.}
\newtheorem{theorem}{\textbf{Theorem}}

\def\cvprPaperID{2048} 
\def\confName{CVPR}
\def\confYear{2023}

\begin{document}

\title{Reliable and Efficient Evaluation of Adversarial Robustness \\ for Deep Hashing-Based Retrieval}

\author{Xunguang Wang\textsuperscript{1},
	Jiawang Bai\textsuperscript{2}, Xinyue Xu\textsuperscript{1}, Xiaomeng Li\textsuperscript{1} \\
	\textsuperscript{1}The Hong Kong University of Science and Technology \\ \textsuperscript{2}Tsinghua University \\
	{\tt\small xunguangwang@gmail.com, bjw19@mails.tsinghua.edu.cn, xxcub@ust.hk, eexmli@ust.hk}
}

\maketitle

\begin{abstract}
   Deep hashing has been extensively applied to massive image retrieval due to its efficiency and effectiveness. Recently, several adversarial attacks have been presented to reveal the vulnerability of deep hashing models against adversarial examples. However, existing attack methods suffer from degraded performance or inefficiency because they underutilize the semantic relations between original samples or spend a lot of time learning these relations with a deep neural network. In this paper, we propose a novel Pharos-guided Attack, dubbed \textbf{PgA}, to evaluate the adversarial robustness of deep hashing networks reliably and efficiently. Specifically, we design \textit{pharos code} to represent the semantics of the benign image, which preserves the similarity to semantically relevant samples and dissimilarity to irrelevant ones. It is proven that we can quickly calculate the pharos code via a simple math formula. Accordingly, PgA can directly conduct a reliable and efficient attack on deep hashing-based retrieval by maximizing the similarity between the hash code of the adversarial example and the pharos code. Extensive experiments on the benchmark datasets verify that the proposed algorithm outperforms the prior state-of-the-arts in both attack strength and speed.
\end{abstract}

\section{Introduction}
It is challenging to rapidly and effectively search for the required information from vast collections in the current era of big data. Learning to hash (hashing) \cite{wang2017survey} has attracted much attention in large-scale image retrieval due to its exceptional benefits in efficient XOR operation and low storage cost by mapping high-dimensional data to compact binary codes. Particularly, deep hashing \cite{xia2014supervised, li2016feature, cao2017hashnet} that learns nonlinear hash functions with deep neural networks (DNNs) has become a predominant image search technique since it delivers better retrieval accuracy than conventional hashing. 

\begin{figure}[t]
\vspace{-0.6cm}
\begin{center}
\includegraphics[width=0.68\columnwidth]{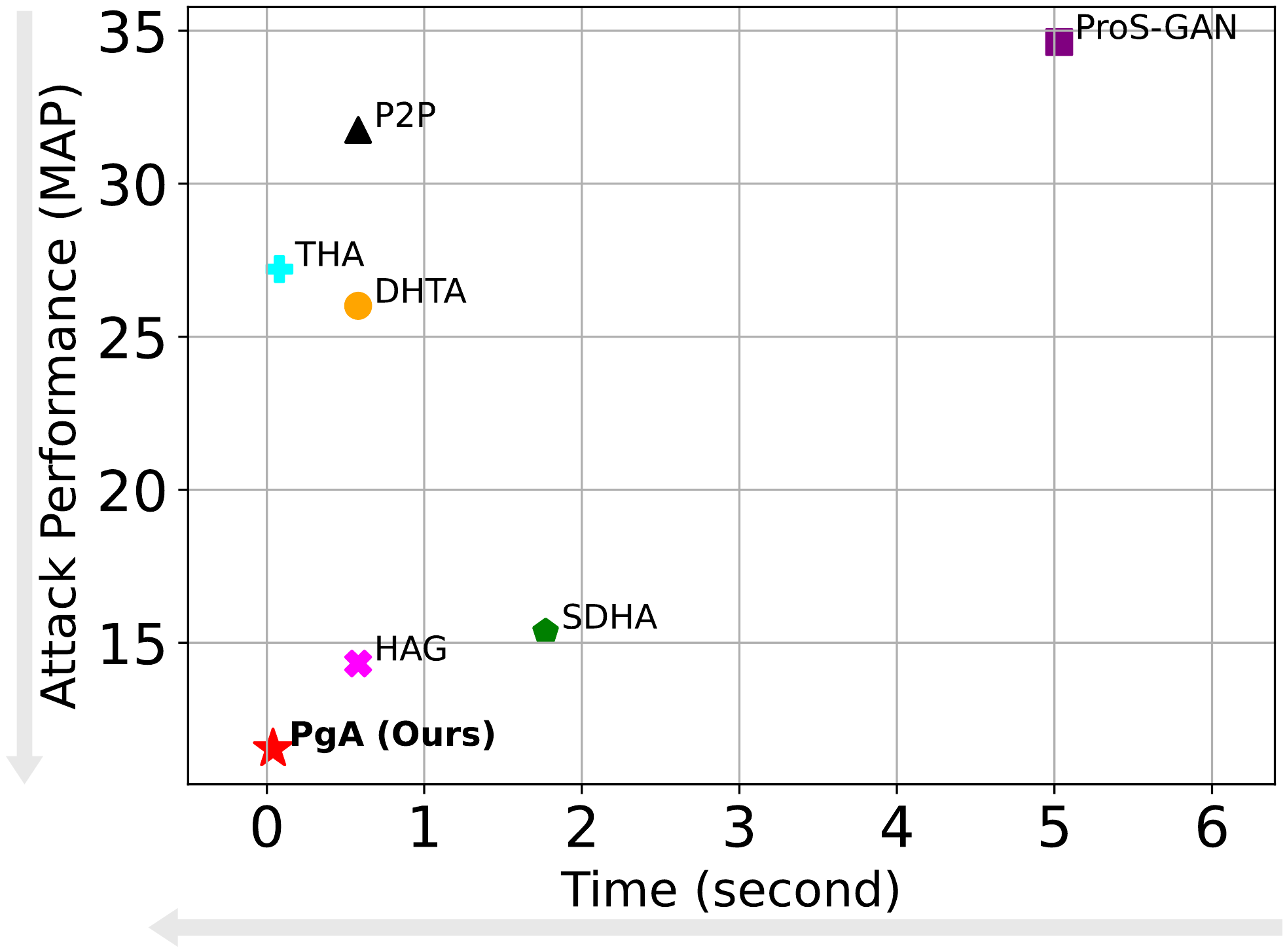}
\end{center}
\vspace{-0.6cm}
\caption{\small Attack performance of deep hashing models against adversarial attacks. We use the retrieval metric, \textit{i.e.}, mean average precision (MAP) \cite{yang2018adversarial} to measure the attack performance. The horizontal axis indicates the average time (seconds) for constructing per adversarial sample. Our method is closest to the origin of coordinates, yielding the highest efficiency.}
\label{fig:motivation}
\vspace{-0.6cm}
\end{figure}

Recent works \cite{yang2018adversarial, bai2020targeted, wang2021prototype, wang2021targeted, zhang2021targeted, xiao2021you, lu2021smart} have revealed that deep hashing models are susceptible to adversarial examples. Although these imperceptible samples are crafted by adding small perturbations to original samples, they are sufficient to deceive models into making inaccurate predictions. There is no doubt that such malicious attacks pose grave security threats to image retrieval systems based on deep hashing. In a deep hashing-based face recognition system, for instance, adversarial examples can mislead the system into matching the faces of specific individuals in the database, infiltrating the system effectively. Consequently, there is significant demand for research into these security concerns in deep hashing-based retrieval.

Some studies \cite{yang2018adversarial,bai2020targeted,wang2021prototype,wang2021targeted,lu2021smart,chen2022adversarial} have been conducted on adversarial attacks and adversarial defenses in deep hashing-based retrieval at present. However, there is no reliable and efficient attack algorithm as a benchmark to evaluate the defense performance against adversarial attacks in hashing-based retrieval. A strong adversarial attack method with high efficiency can provide reasonable benchmarks of model robustness and facilitate the development of adversarial defense strategies. Although existing attack techniques have revealed the vulnerability of deep hashing networks in handling adversarial samples, they are neither reliable nor efficient in evaluating the adversarial robustness (\textit{i.e.}, defense performance) of deep hashing models. Firstly, these methods \cite{yang2018adversarial, bai2020targeted, lu2021smart} suffer from limited attack performances because they do not fully leverage the semantic relevance between available samples. For instance, SDHA \cite{lu2021smart} only reduces the similarity of adversarial samples with their semantically relevant images, ignoring more irrelevant ones. Secondly, though some hashing attack methods simultaneously consider similar and dissimilar pairs, they use time-consuming neural networks to learn discriminative semantic representations from these pairs for the precise semantic attack, \textit{e.g.}, ProS-GAN \cite{wang2021prototype} and THA \cite{wang2021targeted}. In this paper, we focus on improving the deficiencies of previous hashing attacks in both effectiveness and efficiency, as shown in Fig. \ref{fig:motivation}. 


In this study, we propose Pharos-guided Attack (PgA) for reliable and efficient adversarial robustness evaluation of deep hashing networks. The core idea is to quickly compute the pharos code, which reflects the semantics of the original image, and then to use the pharos code to direct the generation of the potent adversarial sample.
Specifically, we first design an optimal hash code (namely \textit{pharos code}) as the discriminative representative of the benign image semantics, which maintains the similarities to hash codes of positive samples and the dissimilarities to those of negative ones (positive and negative samples represent semantically relevant and irrelevant samples of the benign image, respectively).
Benefiting from the binary property of hash codes, we prove that the proposed \textit{Pharos Generation Method} (PGM) can directly calculate the pharos code through a simple mathematical formula (refer to Theorem \ref{theo:pgm}). Thus, the pharos codes of the input data are calculated immediately before the adversarial attack. Subsequently, based on the pharos code, it is feasible to carry out an efficient adversarial hashing attack by maximizing the Hamming distance between the hash code of the adversarial example and the pharos code. Due to the excellence of the pharos codes, our attack manner can considerably enhance the effectiveness and efficiency of adversarial robustness verification, as shown in Fig. \ref{fig:motivation}. In summary, our main contributions are as follows:
\begin{itemize}
    \vspace{-0.7em}
    \item We create the pharos code as the precise semantic representative of the original image content to aid in the construction of the adversarial attack framework for deep hashing-based retrieval. It should be emphasized that our proven mathematical formula in PGM can generate the pharos code instantly.
    \vspace{-0.7em}
    \item A simple pharos-guided attack algorithm is provided, \textit{i.e.}, PgA, which is a reliable and efficient method to evaluate the adversarial robustness of deep hashing networks.
    \vspace{-0.7em}
    \item Extensive experiments demonstrate that PgA can be applied to deep hashing frameworks and achieve state-of-the-art attack performance with high efficiency.
\end{itemize}

\section{Related Work}
\subsection{Deep Hashing based Image Retrieval}
With the remarkable success of deep learning on many vision tasks \cite{
he2016deep,girshick2015fast,tang2004video,deng2019mutual,wang2018cosface}, deep hashing methods have been well developed for large-scale image retrieval, yielding superior performances than the traditional hashing methods based on hand-crafted features. The pioneering CNNH \cite{xia2014supervised} adopts a two-stage strategy, \textit{i.e.}, hash code generation of training data and hash function construction with DNN. 
Recently, deep hashing methods \cite{lai2015simultaneous, zhu2016deep, li2016feature, liu2016deep, li2017deep, cao2017hashnet, jiang2017asymmetric, cao2018deep, su2018greedy, wang2020deep, doan2022one} focus on joint feature learning and hash code encoding into an end-to-end DNN for the better quality of hash codes. A notable work is DPSH \cite{li2016feature}, which simultaneously learns the visual features of data points and preserves their semantic similarity with a pairwise-similarity loss. To alleviate data imbalance between positive and negative pairs, HashNet \cite{cao2017hashnet} adopts a weighted strategy in the pairwise loss function. 
Different from the pairwise similarity learning, CSQ \cite{yuan2020central} can generate high-quality hash codes by enforcing them close to pre-defined hash centers.

\subsection{Adversarial Attack}
In image classification, numerous \textit{adversarial attack} methods \cite{szegedy2013intriguing,goodfellow2014explaining,kurakin2016adversarial,moosavi2016deepfool,madry2017towards,carlini2017towards,dong2018boosting,papernot2017practical,chen2017zoo,ilyas2018black} have been developed to fool the well-trained classifiers by constructing adversarial examples, since the intriguing properties \cite{szegedy2013intriguing, biggio2013evasion} of adversarial samples are discovered. For example, FGSM \cite{goodfellow2014explaining} crafts adversarial samples by maximizing the loss along the gradient direction with a large step. As the multi-step variant of FGSM, I-FGSM \cite{kurakin2016adversarial} and PGD \cite{madry2017towards} iteratively update perturbations with small steps for better attack performance. 

Recently, researchers have extended adversarial attacks to deep hashing-based image retrieval \cite{yang2018adversarial,bai2020targeted,wang2021prototype,wang2021targeted,lu2021smart,zhang2021targeted,bai2022practical,wang2023cgat}. Existing adversarial attack methods for deep hashing can be organized into two categories: \textit{{non-targeted attack}} and \textit{{targeted attack}}. For a non-targeted attack in hashing-based retrieval, its goal is to generate adversarial examples that can confuse the hashing model to retrieve results irrelevant to the original image \cite{bai2020targeted,wang2021prototype}. Achieving the non-targeted attack by minimizing the hash code similarity between the adversarial example and the original sample, Yang \textit{et al.} \cite{yang2018adversarial} proposed \textbf{HAG}, the first adversarial attack method on deep hashing.
\textbf{SDHA} \cite{lu2021smart} generates more effective adversarial queries due to staying away from the relevant images of the benign sample, while HAG only takes the original image into consideration.
As for the targeted attack, it aims to construct adversarial examples whose retrieved images are semantically relevant to the given target label \cite{bai2020targeted,wang2021prototype}. To achieve the targeted attack, \textbf{P2P} and \textbf{DHTA} \cite{bai2020targeted} obtain the anchor code as the representative of the target label to direct the generation of the adversarial sample. Subsequently, Wang \textit{et al.} \cite{wang2021targeted} defined the prototype code as the target code to reach a better targeted attack, which is called \textbf{THA} in this paper. \textbf{ProS-GAN} \cite{wang2021prototype} designs a generative framework for efficient targeted hashing attack under the test phase. Different from the above white-box scenarios, Xiao \textit{et al.} \cite{xiao2021you} proposed the targeted black-box attack {NAG} by enhancing the transferability of adversarial examples.


Unlike the prior work \cite{bai2020targeted} where the anchor code is obtained by a few instances with the same semantics, we propose the pharos code, which preserves the semantic similarity to hash codes of relevant samples and dissimilarity to those of irrelevant samples.
Moreover, we use the proven mathematical formula (\textit{i.e.}, PGM) to instantly calculate the pharos code before the adversarial attack, instead of learning the prototype codes \cite{wang2021prototype,wang2021targeted} through time-consuming neural networks as ProS-GAN and THA do. Hence, our pharos code is more suited for efficient adversarial robustness evaluation of deep hashing models.

\subsection{Adversarial Training}
\textit{Adversarial training} \cite{goodfellow2014explaining, madry2017towards} aims to augment the training data with generated adversarial examples, which is the most robust training strategy against various adversarial attacks. Thus, modifications \cite{zhang2019theoretically, wong2020fast, pang2020bag} and applications \cite{li2021divergence, utrera2020adversarially,bai2022improving} of adversarial training have emerged to improve the robustness and generalization of DNNs.
For deep hashing-based retrieval, \cite{wang2021targeted} proposed the first effective adversarial training algorithm based on the targeted attack (dubbed \textbf{ATRDH} here) by narrowing the semantic gap between the adversarial samples and the original samples in the Hamming space.

\section{Method}
\subsection{Preliminaries}
We consider that an attacked hashing model $F$ learns from a training set of $N$ data points $O=\{(\boldsymbol{x}_i,\boldsymbol{y}_i)\}_{i=1}^N$, where $\boldsymbol{x}_i$ indicates $i$-th image, and $\boldsymbol{y}_i=[y_{i1},y_{i2},...,y_{iC}]\in \{0,1\}^C$ denotes a label vector of $\boldsymbol{x}_i$. $C$ indicates the total number of classes in the dataset. $y_{ij}=1$ means that $\boldsymbol{x}_i$ belongs to the $j$-th class. If $\boldsymbol{x}_i$ and $\boldsymbol{x}_j$ share at least one common label, they are semantically similar, \textit{i.e.}, $\boldsymbol{x}_j$ is the positive sample of $\boldsymbol{x}_i$. Otherwise, they are semantically dissimilar and $\boldsymbol{x}_j$ is the negative sample of $\boldsymbol{x}_i$.

Deep hashing aims at employing DNNs to transform high-dimensional data into compact binary codes and simultaneously preserves their semantic similarities. For the given hashing model $F$, the hash code $\boldsymbol{b}_i$ of the instance $\boldsymbol{x}_i$ is generated as:
\begin{equation}
    \begin{aligned}
        \boldsymbol{b}_i = {F}(\boldsymbol{x}_i) &= \operatorname{sign}(\boldsymbol{h}_i)= \operatorname{sign}(f_\theta(\boldsymbol{x}_i)),\\ 
        s.t.~ & \boldsymbol{b}_i \in\{-1,1\}^K,
    \end{aligned}
\end{equation}
where $K$ represents the hash code length, and $f(\cdot)$ with parameter $\theta$ is a DNN to
approximate hash function ${F}(\cdot)$. The final binary code $\boldsymbol{b}_i$ is obtained by applying the $\operatorname{sign}(\cdot)$ on the output $\boldsymbol{h}_i$ of $f_\theta(\boldsymbol{x}_i)$. Typically, $f(\cdot)$ is implemented by a convolutional neural network (CNN) and adopts the $\operatorname{tanh}$ activation to simulate sign function at the output layer. 

\subsection{The Proposed Pharos-guided Attack}
\subsubsection{Problem Formulation}
In hashing based retrieval, the goal of adversarial attack (\textit{i.e.}, non-targeted attack) is to craft an adversarial example whose retrieval results are irrelevant to the original sample contents. For credibility, this objective can be achieved by maximizing the hash code distance between the adversarial example and its semantically relevant samples, and simultaneously minimizing the distance from irrelevant samples, rather than the only benign sample. Thus, for a given clean image $\boldsymbol{x}$, the objective of its adversarial example $\boldsymbol{x}^\prime$ is formulated as follows:
\begin{equation}
	\begin{aligned}
	    \max_{\boldsymbol{x}^\prime} \!\sum_{i}^{N_{\rm{p}}}\!\sum_{j}^{N_{\rm{n}}} [w_i {D}( {F}(\boldsymbol{x}^\prime), &{F}(\boldsymbol{x}_i^{(\rm{p})}))\!-\! w_j {D}({F}(\boldsymbol{x}^\prime), \!{F}(\boldsymbol{x}_j^{(\rm{n})}))], \\
     s.t.~ \|\boldsymbol{x} &- \boldsymbol{x}^\prime\|_p \leq \epsilon,
	\end{aligned}
\label{eq:task}
\end{equation}
where $F(\cdot)$ is the hashing function approximated by the deep model $f(\cdot)$, and $D(\cdot, \cdot)$ is a distance metric. $w_i$ and $w_j$ represent distance weights. $\boldsymbol{x}_i^{(\rm{p})}$ is a positive sample semantically related to the original sample $\boldsymbol{x}$, and $\boldsymbol{x}_j^{(\rm{n})}$ is a negative sample of $\boldsymbol{x}$. Because this maximizing term of Eq. (\ref{eq:task}) can push the hash code of the adversarial example close to those of unrelated samples and away from semantically relevant samples, optimal attack strength would come true in theory. $N_{\rm{p}}$ and $N_{\rm{n}}$ are the number of the positive samples and the negative samples, respectively. $\|\cdot\|_p$ ($p={1,2,\infty}$) is $L_p$ norm which keeps the pixel difference between the adversarial sample and the original sample no more than $\epsilon$ for the imperceptible property of adversarial perturbations.

\subsubsection{Generation of Pharos Codes}
Actually, the maximized objective in Eq. (\ref{eq:task}) is equivalent to finding a hash code $\boldsymbol{b}^\prime$ in hashing, which satisfies:
\begin{equation}
	\begin{aligned}
	    \max_{\boldsymbol{b}^\prime} \sum_{i}\sum_{j} [ w_i {D}_{\rm{H}}(\boldsymbol{b}^\prime, \boldsymbol{b}_i^{(\rm{p})}) - w_j {D}_{\rm{H}}(\boldsymbol{b}^\prime, \boldsymbol{b}_j^{(\rm{n})}) ]
	\end{aligned},
\label{eq:pro_max_1}
\end{equation}
where $D_{\rm{H}}$ is Hamming distance measure. $\boldsymbol{b}_i^{(\rm{p})}$ is the hash code of the positive sample $\boldsymbol{x}_i^{(\rm{p})}$, and $\boldsymbol{b}_j^{(\rm{n})}$ is the binary code of the negative sample $\boldsymbol{x}_j^{(\rm{n})}$, \textit{i.e.}, $\boldsymbol{b}_i^{(\rm{p})}=F(\boldsymbol{x}_i^{(\rm{p})}), \boldsymbol{b}_j^{(\rm{n})}=F(\boldsymbol{x}_j^{(\rm{n})})$. Subsequently, we can optimize the adversarial example by minimizing the distance from $\boldsymbol{b}^\prime$, \textit{i.e.},
\begin{equation}
    \min_{\boldsymbol{x}^\prime} D_{\rm{H}} (\boldsymbol{b}^\prime, F(\boldsymbol{x}^\prime)).
    \label{eq:pro_max_2}
\end{equation}
For any hash code $\boldsymbol{\hat{b}}$ and $\boldsymbol{\check{b}}$, we know that $D_{\rm{H}}(\boldsymbol{\hat{b}}, \boldsymbol{\check{b}})=\frac{1}{2}(K-\boldsymbol{\hat{b}}^{\top}\boldsymbol{\check{b}})$. Accordingly, we deduce that $D_{\rm{H}}(\boldsymbol{\hat{b}}, \boldsymbol{\check{b}}) = K - D_{\rm{H}}(-\boldsymbol{\hat{b}}, \boldsymbol{\check{b}})$. Let hash code $\boldsymbol{b}^\star=-\boldsymbol{b}^\prime$, the Eq. (\ref{eq:pro_max_1}) and (\ref{eq:pro_max_2}) can be reformulated as follows:
\begin{equation}
	\begin{aligned}
	    \min_{\boldsymbol{b}^\star} \!\sum_{i}\!\sum_{j}\{ w_i [{D}_{\rm{H}}(\boldsymbol{b}^\star, \boldsymbol{b}_i^{(\rm{p})})\!-\!K] &\!-\! w_j [{D}_{\rm{H}}(\boldsymbol{b}^\star, \boldsymbol{b}_j^{(\rm{n})})\!-\!K]\}, \\
	    \max_{\boldsymbol{x}^\prime} D_{\rm{H}} (\boldsymbol{b}^\star, F(\boldsymbol{x}^\prime)) &- K.
	\end{aligned}
\label{eq:pro_min}
\end{equation}
Removing the constants, the Eq. (\ref{eq:pro_min}) can be written as:
\begin{equation}
	\begin{aligned}
	    \min_{\boldsymbol{b}^\star} \sum_{i}\sum_{j} [w_i {D}_{\rm{H}}(\boldsymbol{b}^\star, \boldsymbol{b}_i^{(\rm{p})}) &- w_j {D}_{\rm{H}}(\boldsymbol{b}^\star, \boldsymbol{b}_j^{(\rm{n})})], \\
	    \max_{\boldsymbol{x}^\prime} D_{\rm{H}} (\boldsymbol{b}^\star &, F(\boldsymbol{x}^\prime)).
	\end{aligned}
\label{eq:obj}
\end{equation}
Due to the binary characteristic of the hash code, we can directly calculate the optimal code (named \textit{pharos code} $\boldsymbol{b}^\star$) in the problem (\ref{eq:obj}) by a simple mathematical formula, as shown in Theorem \ref{theo:pgm}.
\begin{theorem}
\label{theo:pgm}
Pharos code $\boldsymbol{b}^\star$ formulated in Eq. (\ref{eq:obj}) can be calculated by the sum of the difference between the weighted positive hash codes and those of negative ones, \textit{i.e.},
\begin{equation}
	\begin{aligned}
	    \boldsymbol{b}^\star = \operatorname{sign}(\sum_{i}^{N_{\rm{p}}}\sum_{j}^{N_{\rm{n}}}(w_{i}\boldsymbol{b}_i^{(\rm{p})} - w_{j}\boldsymbol{b}_j^{(\rm{n})}) )
	\end{aligned},
\label{eq:cen}
\end{equation}
\end{theorem}
We name the way of obtaining the pharos code with Eq. (\ref{eq:cen}) as Pharos Generation Method (PGM). $\operatorname{sign}(\cdot)$ in Theorem \ref{theo:pgm} is the sign function. The proof of PGM is shown in \cref{a1}. In addition, we define the $w_i$ and $w_j$ as follows:
\begin{equation}
    w_i = s_i, \quad w_j = 1-s_j,
\end{equation}
where $s_{i/j}$ ($s_{i/j}\in [0,1]$) denotes the similarity between the adversarial example and the $i/j$-th benign sample. If labels $\boldsymbol{y}_i$ and $\boldsymbol{y}_j$ of $\boldsymbol{x}_i$ and $\boldsymbol{x}_j$ are given, we can calculate $s_{i/j}$ by Dice coefficient, \textit{i.e.}, $s_{i/j}=\frac{2\vert\boldsymbol{y}\cap \boldsymbol{y}_{i/j} \vert}{\vert\boldsymbol{y}\vert+\vert\boldsymbol{y}_{i/j}\vert}$. Otherwise, $s_{i/j}$ is usually determined by the optimization objective of the attacked hashing model. For instance, $s_i=1$ and $s_j=0$ are widely adopted in learning to hash.

\subsubsection{Generating Adversarial Examples}
Since the pharos code is found, the attack problem described in Eq. (\ref{eq:task}) can be translated into the following objective under the $L_{\infty}$ constraint:
\begin{equation}
    \begin{aligned}
        \max_{\boldsymbol{x}^\prime} D_{\rm{H}} (\boldsymbol{b}^\star, F(\boldsymbol{x}^\prime)), \quad
        s.t.~ \|\boldsymbol{x} & - \boldsymbol{x}^\prime\|_{\infty} \leq \epsilon
    \end{aligned}.
    \label{eq:att}
\end{equation}
According to $D_{\rm{H}}(\boldsymbol{\hat{b}}, \boldsymbol{\check{b}})=\frac{1}{2}(K-\boldsymbol{\hat{b}}^{\top}\boldsymbol{\check{b}})$, Eq. (\ref{eq:att}) is equivalent to:
\begin{equation}
    \begin{aligned}
        \max_{\boldsymbol{x}^\prime} ~\mathcal{L}_{a}= -\frac{1}{K}(\boldsymbol{b}^\star)^{\top} f_{\theta}(\boldsymbol{x}^\prime),~
        s.t.~ \|\boldsymbol{x} - \boldsymbol{x}^\prime\|_{\infty} \leq \epsilon
    \end{aligned}.
    \label{eq:pga}
\end{equation}
However, Eq. (\ref{eq:pga}) focuses on the sum similarity between $f_{\theta}(\boldsymbol{x}^\prime)$ and $\boldsymbol{b}^\star$, and can not effectively promote that each bit in $f_{\theta}(\boldsymbol{x}^\prime)$ differs in sign from $\boldsymbol{b}^\star$. Intuitively, bits with small differences between $f_{\theta}(\boldsymbol{x}^\prime)$ and $\boldsymbol{b}^\star$ should be given big weights. Hence, we add an weighting vector $\boldsymbol{\omega}$ on $\mathcal{L}_{a}$ to enforce each bit of $f_{\theta}(\boldsymbol{x}^\prime)$ away from those bits of $\boldsymbol{b}^\star$. Formally, 

\begin{figure}[t]
\begin{center}
\includegraphics[width=0.8\columnwidth]{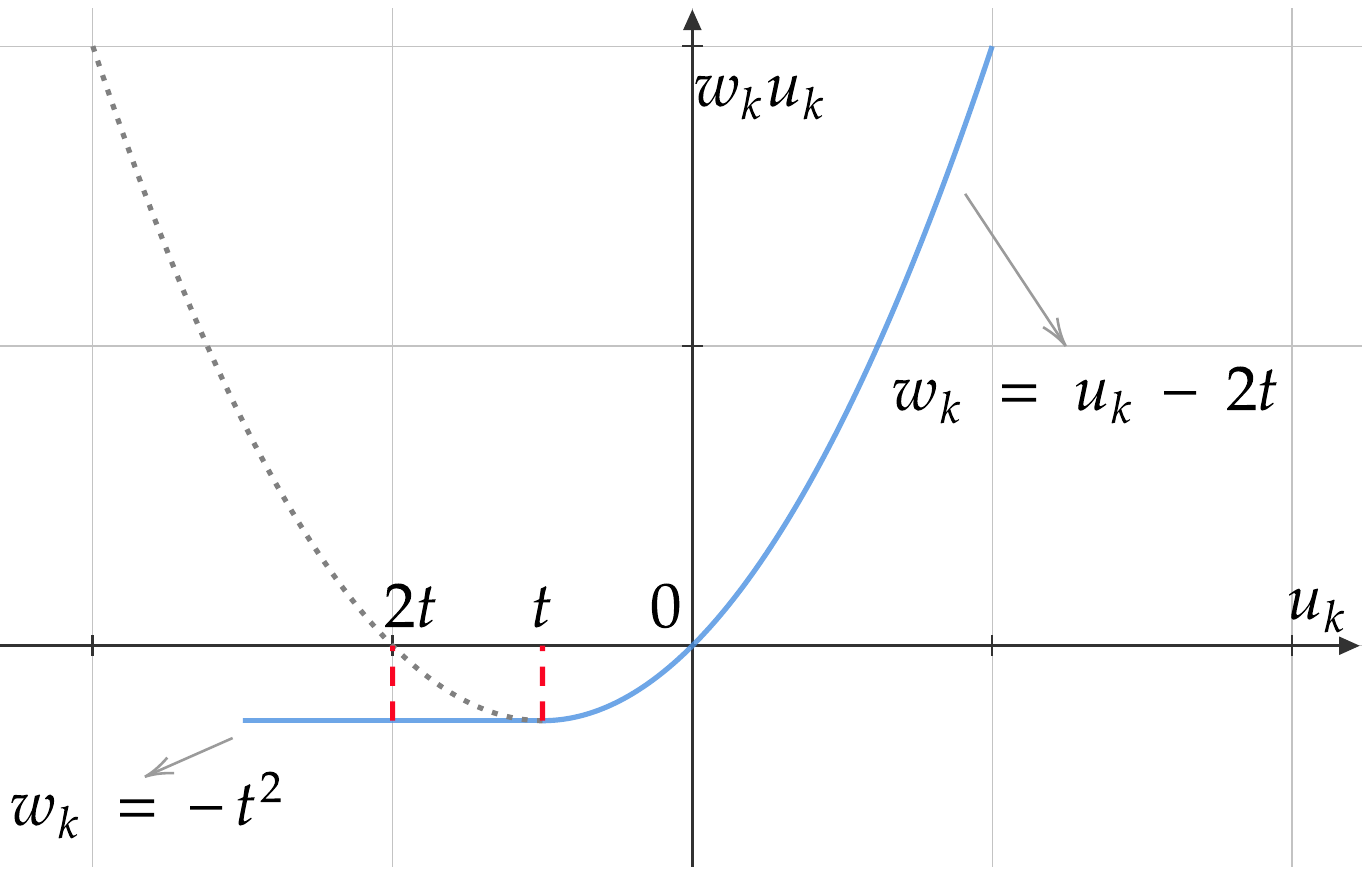}
\end{center}
\vspace{-0.3cm}
\caption{\small The curve of $\omega_k$.}
\label{fig:omega}
\vspace{-0.3cm}
\end{figure}

\begin{equation}
    \begin{aligned}
        \mathcal{L}_{a}= -\frac{1}{K} \boldsymbol{\omega}^{\top}\boldsymbol{u}, \quad \boldsymbol{u}= \boldsymbol{b}^\star \circ f_{\theta}(\boldsymbol{x}^\prime)
    \end{aligned},
    \label{eq:ada_pga}
\end{equation}
where $\circ$ represents Hadamard product, and $\boldsymbol{\omega}$ has the same dimensions as $\boldsymbol{b}^\star$. For efficiency, it is not necessary to enforce each element $u_k$ of $\boldsymbol{u}$ to approximate $-1$. In this case, we set $u_k$ to converge at $t$ ($-1<t<0$). Thus, the component $\boldsymbol{\omega}_{k}$ of $\boldsymbol{\omega}$ is defined as
\begin{equation}
    \begin{aligned}
        \omega_{k}=\left\{\begin{array}{ll}
             u_k-2t, \quad & u_k > t  \\
             -t^2, \quad & \text{otherwise}
        \end{array}\right.
    \end{aligned},
\end{equation}
where $u_k$ is the $k$-th element of $\boldsymbol{u}$. As shown in Figure \ref{fig:omega}, $t$ controls the margin between ${b}^\star_k$ and $ f_{\theta}(\boldsymbol{x}^\prime)_k$. $t$ is set to $-0.8$ by default.
Furthermore, we follow \cite{yang2018adversarial} to make our pharos-guided attack focus on different sign between $f(\boldsymbol{x}^\prime)$ and $\boldsymbol{b}^\star$ for efficiency, \textit{i.e.},
\begin{equation}
    \begin{aligned}
        \mathcal{L}_{a}= -\frac{1}{\pi} [\boldsymbol{m}\circ\boldsymbol{\omega}]^{\top}\boldsymbol{u}, \quad \boldsymbol{u}= \boldsymbol{b}^\star \circ f_{\theta}(\boldsymbol{x}^\prime)
    \end{aligned},
    \label{eq:eff_pga}
\end{equation}
where $\boldsymbol{m}\in \{0,1\}^K$ is a mask and $\pi$ is the number of non-zero elements in $\boldsymbol{m}$. The element $m_{k}$ of $\boldsymbol{m}$ is defined as
\begin{equation}
    \begin{aligned}
        m_{k}=\left\{\begin{array}{ll}
             1, \quad & u_k > t  \\
             0, \quad & \text{otherwise}
        \end{array}\right.
    \end{aligned}.
    \label{eq:mask}
\end{equation}
Notably, the pharos-guided attack with Eq. (\ref{eq:pga}) is called \textbf{PgA$^\dagger$}, and that with Eq. (\ref{eq:eff_pga}) is the default \textbf{PgA}.
Unlike HAG and SDHA using SGD \cite{robbins1951stochastic} or Adam \cite{kingma2014adam} optimizer \cite{yang2018adversarial,lu2021smart} with quantities of iterations, this paper adopts PGD \cite{madry2017towards} to optimize $\boldsymbol{x}^\prime$ with $T$ ($T=100$ by default) iterations for efficiency, \textit{i.e.},
\begin{equation}
    \begin{aligned}
        \boldsymbol{x}_T^\prime = {\mathcal{S}}_{\epsilon}(\boldsymbol{x}_{T-1}^\prime + \eta \cdot \operatorname{sign}(\Delta_{\boldsymbol{x}_{T-1}^\prime}\mathcal{L}_{a})),
		~ \boldsymbol{x}^{\prime}_0 = \boldsymbol{x}+\boldsymbol{r},
    \end{aligned}
\end{equation}
where $\eta$ is the step size, and $\mathcal{S}_{\epsilon}$ project $\boldsymbol{x}^\prime$ into the $\epsilon$-ball \cite{madry2017towards} of $\boldsymbol{x}$. $\boldsymbol{r}$ is random noise, sampled from uniform $U(-\epsilon,\epsilon)$.

\begin{table*}[t]
\scriptsize
\begin{center}
\caption{\small MAP (\%) of different attack methods on deep hashing models without defense.}
\vspace{-.3cm}
\label{table:map_hash}
\resizebox{0.8\textwidth}{!}{
\begin{tabular}{lccccccccc}
\toprule
~ & \multicolumn{3}{c}{FLICKR-25K}& \multicolumn{3}{c}{NUS-WIDE}& \multicolumn{3}{c}{MS-COCO} \\
\cmidrule(r){2-4} \cmidrule(r){5-7} \cmidrule{8-10}
Method &16 bits &32 bits &64 bits &16 bits &32 bits &64 bits &16 bits &32 bits &64 bits \\ \midrule
Clean &80.39 &81.35 &91.91 &74.99 &76.72 &77.84 &56.53 &57.93 &57.92 \\
P2P \cite{bai2020targeted} &41.68  &42.69  &41.18  &32.56  &31.76  &31.74  &21.99  &21.78  &21.57   \\
DHTA \cite{bai2020targeted} &34.74 &32.63 &34.58 &26.68 &26.01 &26.33 &19.41 &19.23 &18.05 \\
ProS-GAN \cite{wang2021prototype} &67.29 &76.31 &82.51 &30.50 &34.63 &64.43 &52.12 &53.67 &53.38  \\
THA \cite{wang2021targeted} &38.28 &37.54 &35.67 &30.72 &27.21 &24.27 &24.62 &20.65 &22.71 \\
HAG \cite{yang2018adversarial} &24.75 &24.37 &23.44 &14.55 &14.32 &14.07 &13.91 &13.59 &14.62  \\
SDHA \cite{lu2021smart} &20.50 &19.63 &18.98 &17.09 &15.36 &14.87 &12.07 &12.36 &12.94 \\
\midrule
$\text{PgA}^\dagger$ (Ours) &16.03 &15.49 &15.32 &12.30 &11.93 &12.07 &11.36 &10.56 &11.10 \\
PgA (Ours) &\textbf{15.70} &\textbf{15.18} &\textbf{15.02} &\textbf{11.81} &\textbf{11.53} &\textbf{11.69} &\textbf{9.91} &\textbf{9.41} &\textbf{10.26} \\
\bottomrule
\end{tabular}
}
\end{center}
\end{table*}

\begin{table*}[t]
\scriptsize
\begin{center}
\caption{\small MAP (\%) of multiple attack methods after adversarial training by ATRDH \cite{wang2021targeted}.}
\vspace{-.3cm}
\label{table:map_atrdh}
\resizebox{0.8\textwidth}{!}{
\begin{tabular}{lccccccccc}
\toprule
~ & \multicolumn{3}{c}{FLICKR-25K}& \multicolumn{3}{c}{NUS-WIDE}& \multicolumn{3}{c}{MS-COCO} \\
\cmidrule(r){2-4} \cmidrule(r){5-7} \cmidrule{8-10}
Method &16 bits &32 bits &64 bits &16 bits &32 bits &64 bits &16 bits &32 bits &64 bits \\ 
\midrule
Clean &72.75 &73.34 &73.10 &66.42 &65.90 &68.69 &49.23 &50.87 &49.93  \\
P2P \cite{bai2020targeted} &52.84 &54.56 &57.13 &51.23 &56.70 &54.05 &33.55 &33.24 &30.96  \\
DHTA \cite{bai2020targeted} &49.97 &53.26 &54.04 &49.16 &56.50 &51.91 &32.16 &31.62 &28.35 \\
ProS-GAN \cite{wang2021prototype} &73.33 &73.33 &73.89 &67.07 &67.18 &69.20 &49.59 &51.30 &50.61 \\
THA \cite{wang2021targeted} &48.58 &49.94 &50.43 &50.57 &50.92 &50.86 &32.31 &30.52 &29.33 \\
HAG \cite{yang2018adversarial} &41.99 &45.58 &45.81 &42.15 &47.46 &45.19 &26.50 &27.49 &28.73  \\
SDHA \cite{lu2021smart} &37.65 &40.68 &44.61 &42.39 &47.16 &45.86 &28.00 &27.75 &27.72 \\
\midrule
$\text{PgA}^\dagger$ (Ours) &32.78 &34.67 &35.53 &37.86 &42.84 &39.82 &21.89 &21.74 &\textbf{22.21} \\
PgA (Ours) &\textbf{31.52} &\textbf{33.80} &\textbf{34.52} &\textbf{37.48} &\textbf{42.22} &\textbf{39.54} &\textbf{21.52} &\textbf{21.42} &22.45 \\
\bottomrule
\end{tabular}
}
\end{center}
\vspace{-.6cm}
\end{table*}

\section{Experiments}
\subsection{Experimental Setup}
\textbf{Datasets.}
We adopt three popular datasets used in hashing-based retrieval to evaluate our defense method in extensive experiments: \textit{FLICKR-25K} \cite{huiskes2008mir}, \textit{NUS-WIDE} \cite{chua2009nus} and \textit{MS-COCO} \cite{lin2014microsoft}. The \textbf{FLICKR-25K} dataset comprises 25,000 Flickr images with 38 labels. We sample 1,000 images as the query set and the remaining regarded as the database, following \cite{wang2021targeted}. Moreover, we randomly select 5,000 instances from the database to train hashing models. The \textbf{NUS-WIDE} dataset contains 269,648 images annotated with 81 concepts. We sample a subset with 21 of the most popular concepts, which consists of 195,834 images. 2,100 images are sampled from the subset as queries, while the rest images are regarded as the database. We randomly select 10,500 images from the database for the training set \cite{wang2021prototype}. The \textbf{MS-COCO} dataset consists of 82,783 training samples and 40,504 validation samples, where each instance is annotated with at least one of the 80 categories. After combining the training and the validation set, we randomly pick 5,000 instances from them as queries and the rest as a database. For the training set, 10,000 images are randomly selected from the database. In addition, we make an extra experiment on {CIFAR-10} (refer to \cref{ap:cifar}).

\begin{figure*}[t]
\centering
{\includegraphics[width=0.26\textwidth]{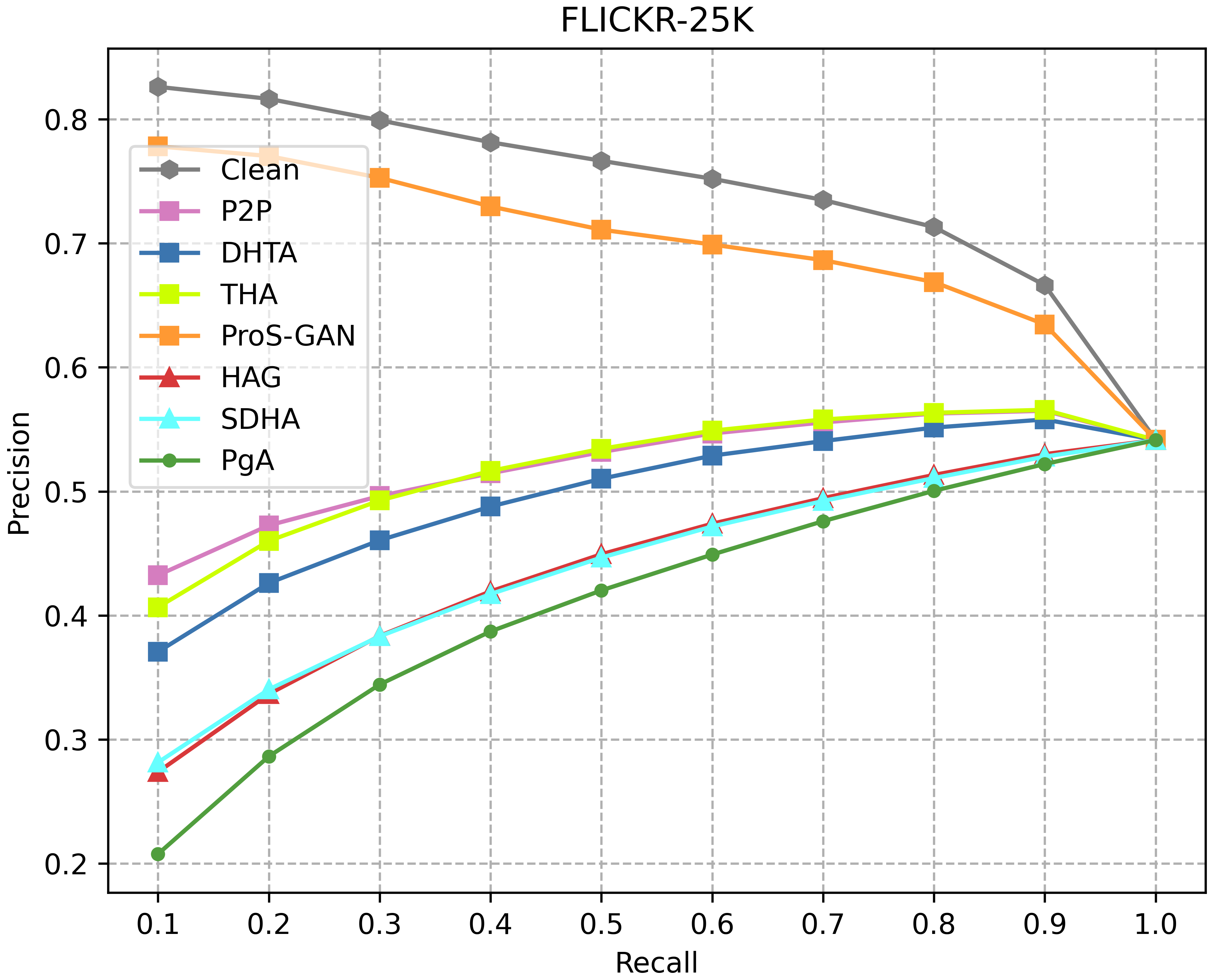}}
\hspace{0.2cm}
{\includegraphics[width=0.26\textwidth]{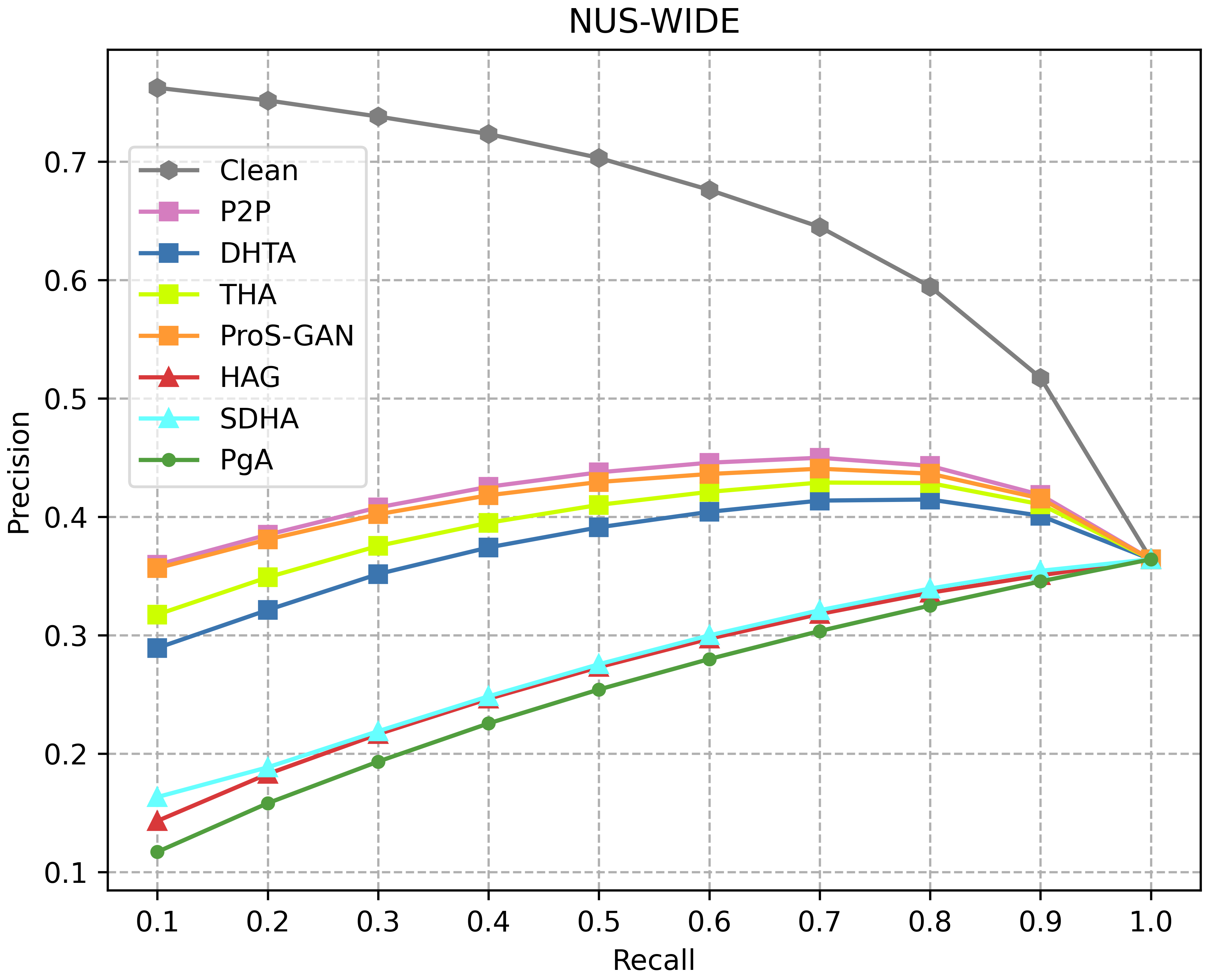}}
\hspace{0.2cm}
{\includegraphics[width=0.26\textwidth]{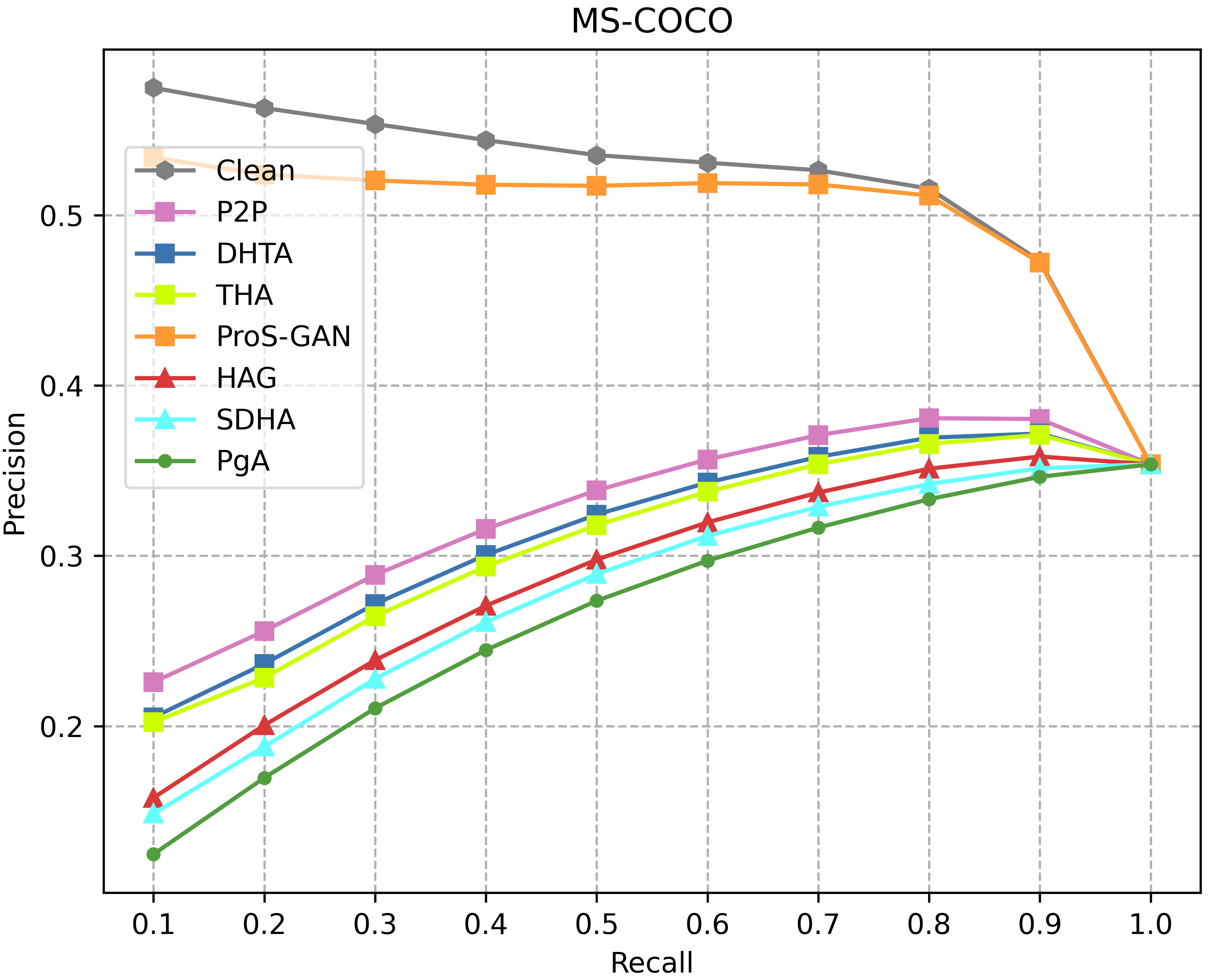}} \\
{\includegraphics[width=0.26\textwidth]{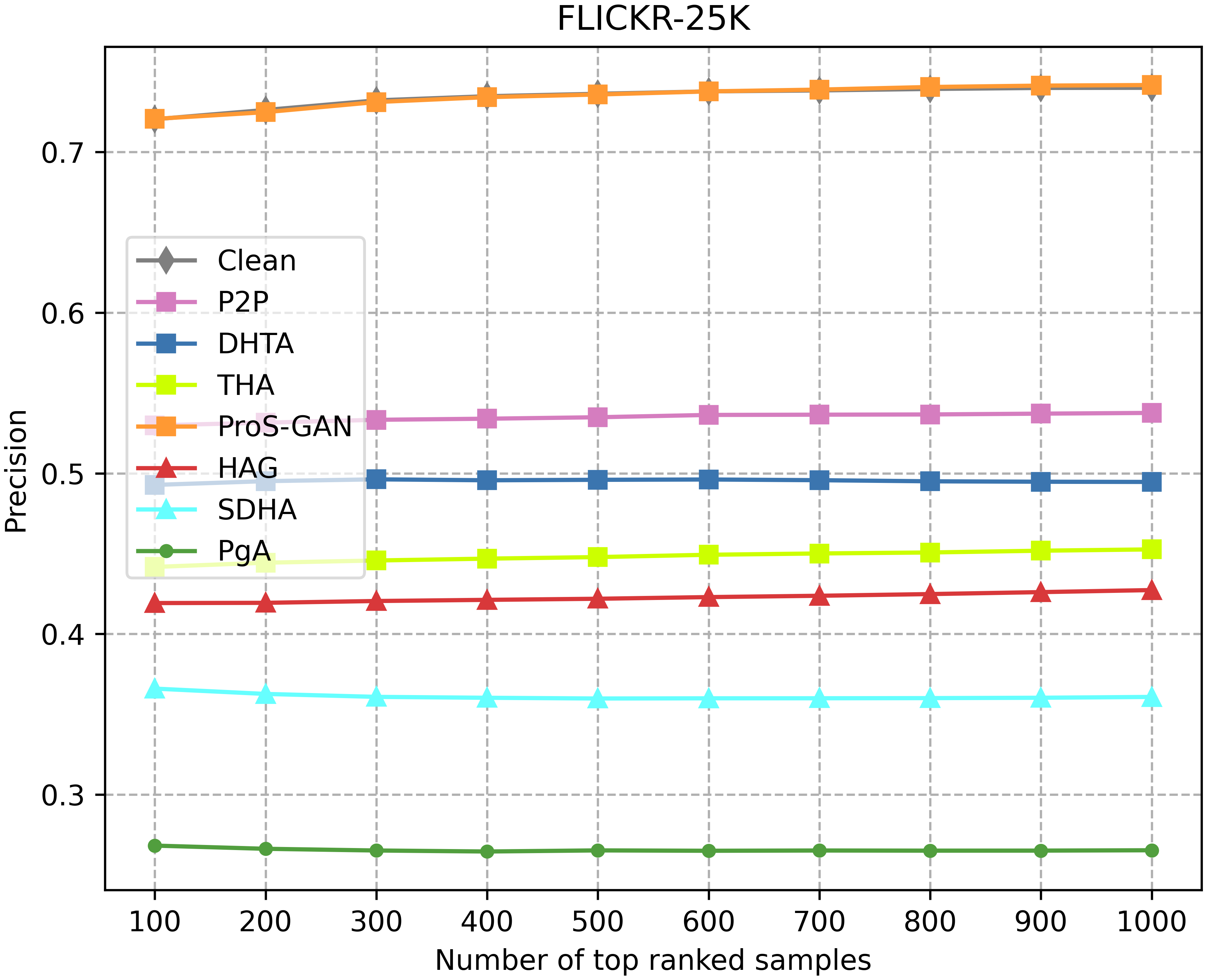}}
\hspace{0.2cm}
{\includegraphics[width=0.26\textwidth]{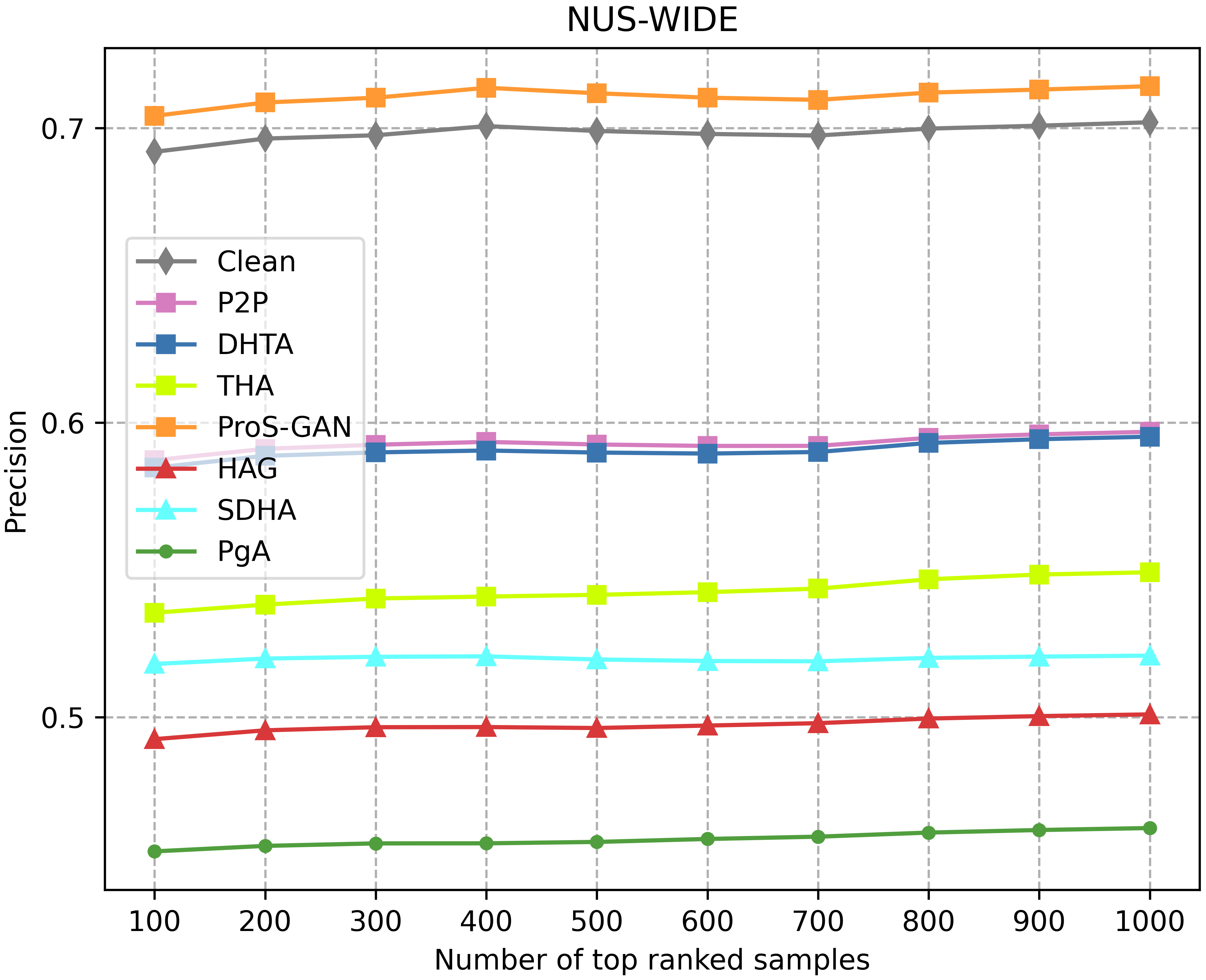}}
\hspace{0.2cm}
{\includegraphics[width=0.26\textwidth]{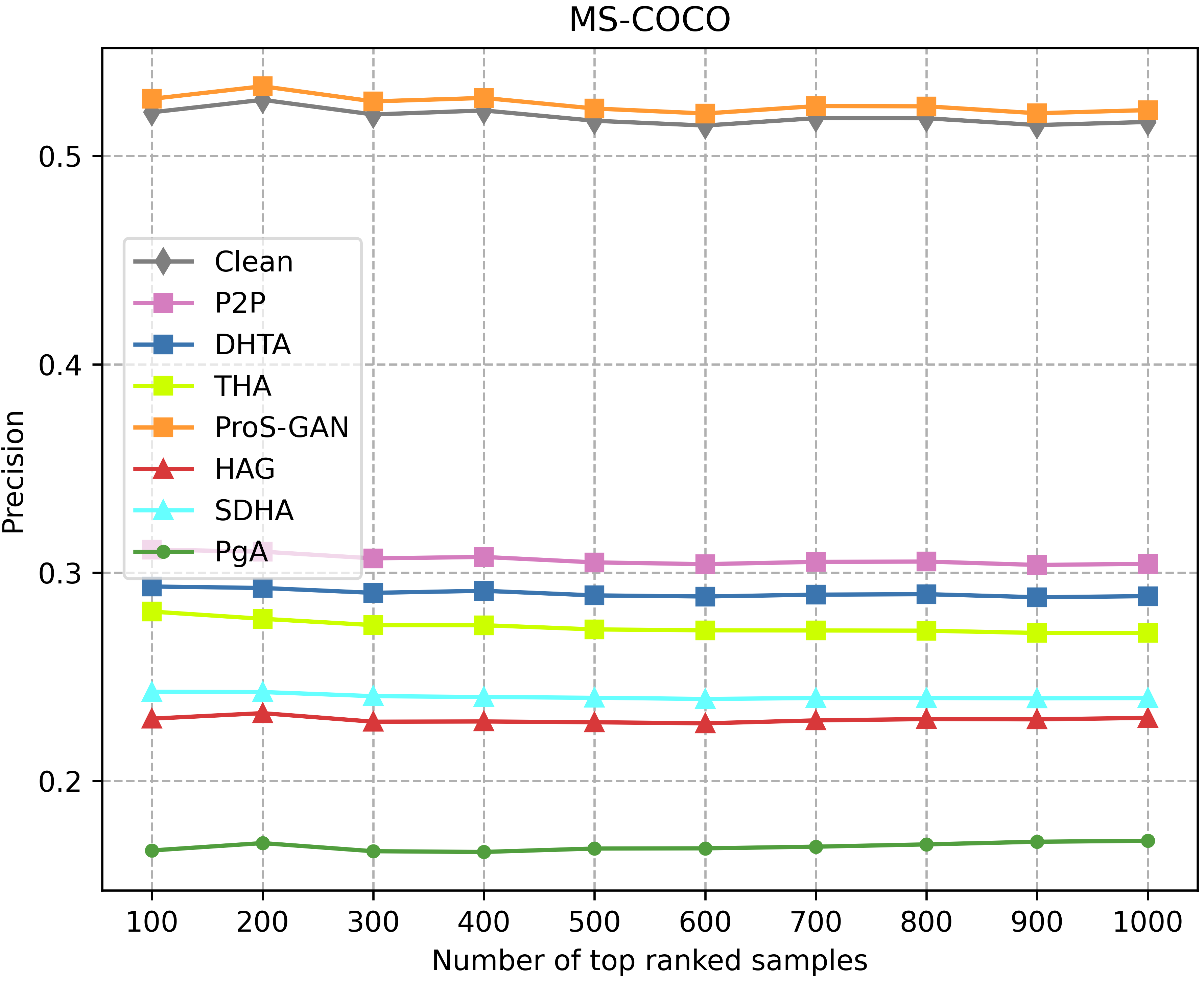}}
\vspace{-.2cm}
\caption{\small Precision-Recall and Precision@topN curves on original deep hashing models with 32 bits code length.}
\label{fig:pr_hash}
\end{figure*}

\begin{figure*}[t]
\vspace{-0.2cm}
\centering
{\includegraphics[width=0.26\textwidth]{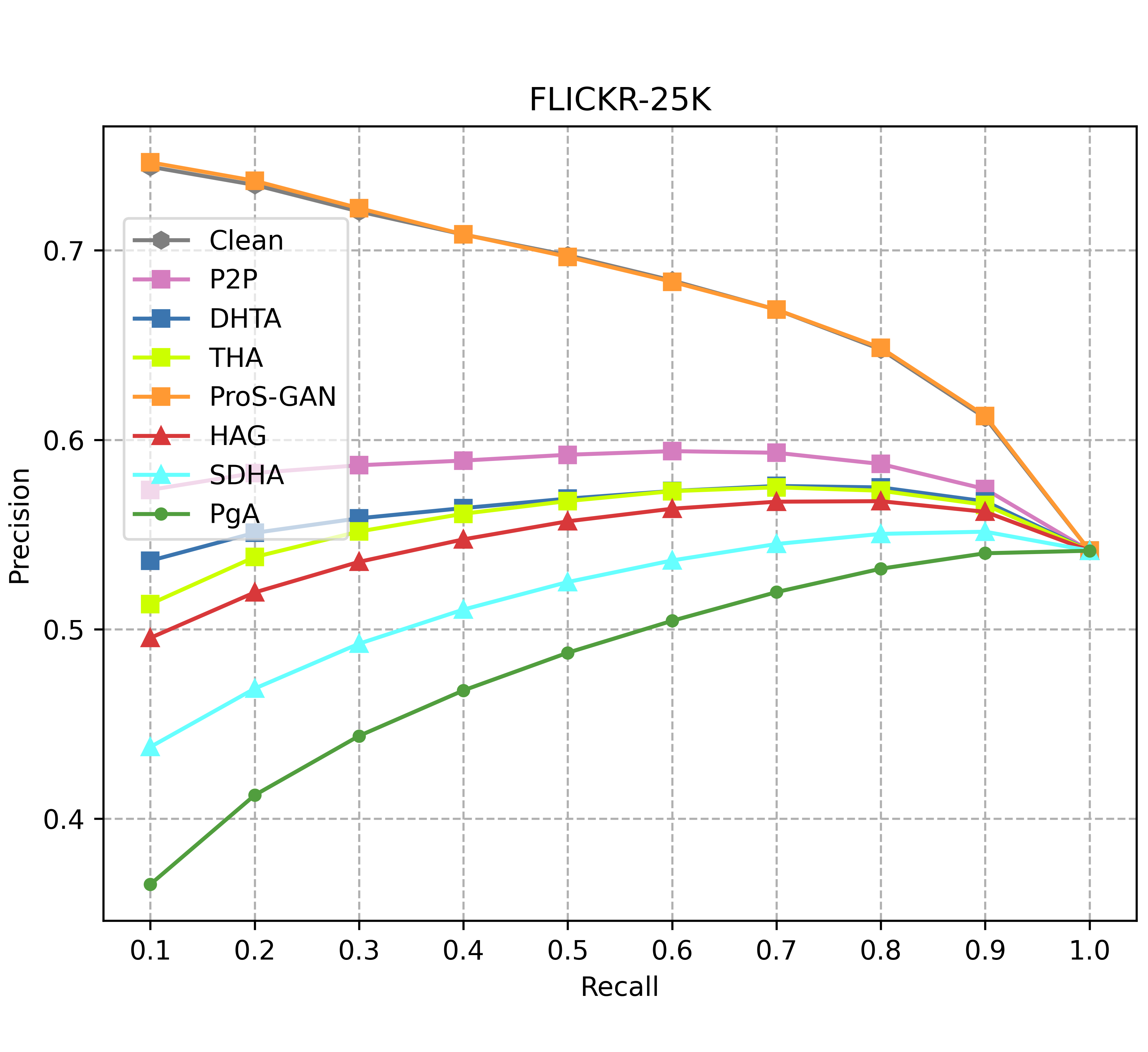}}
\hspace{0.2cm}
{\includegraphics[width=0.26\textwidth]{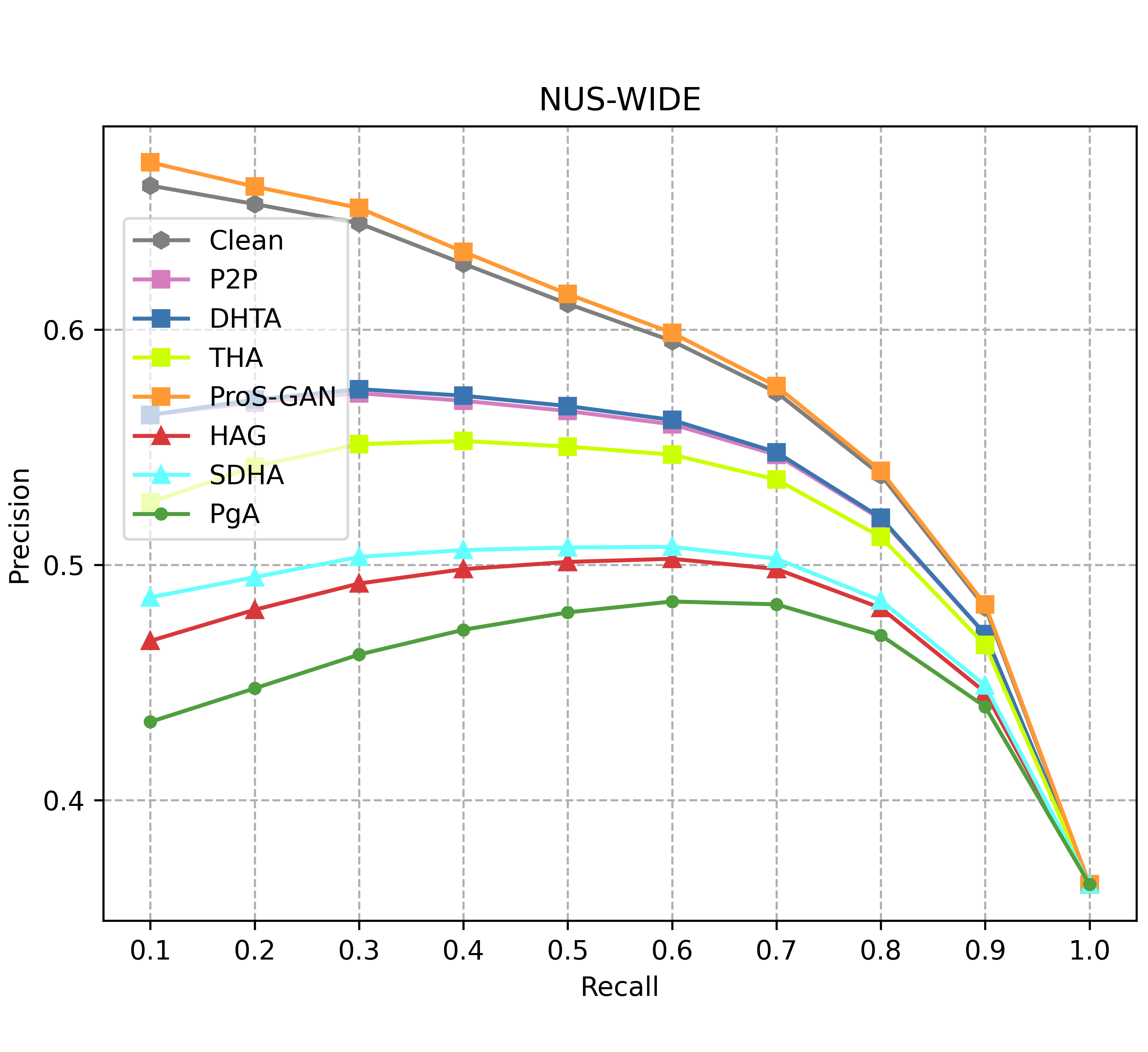}}
\hspace{0.2cm}
{\includegraphics[width=0.26\textwidth]{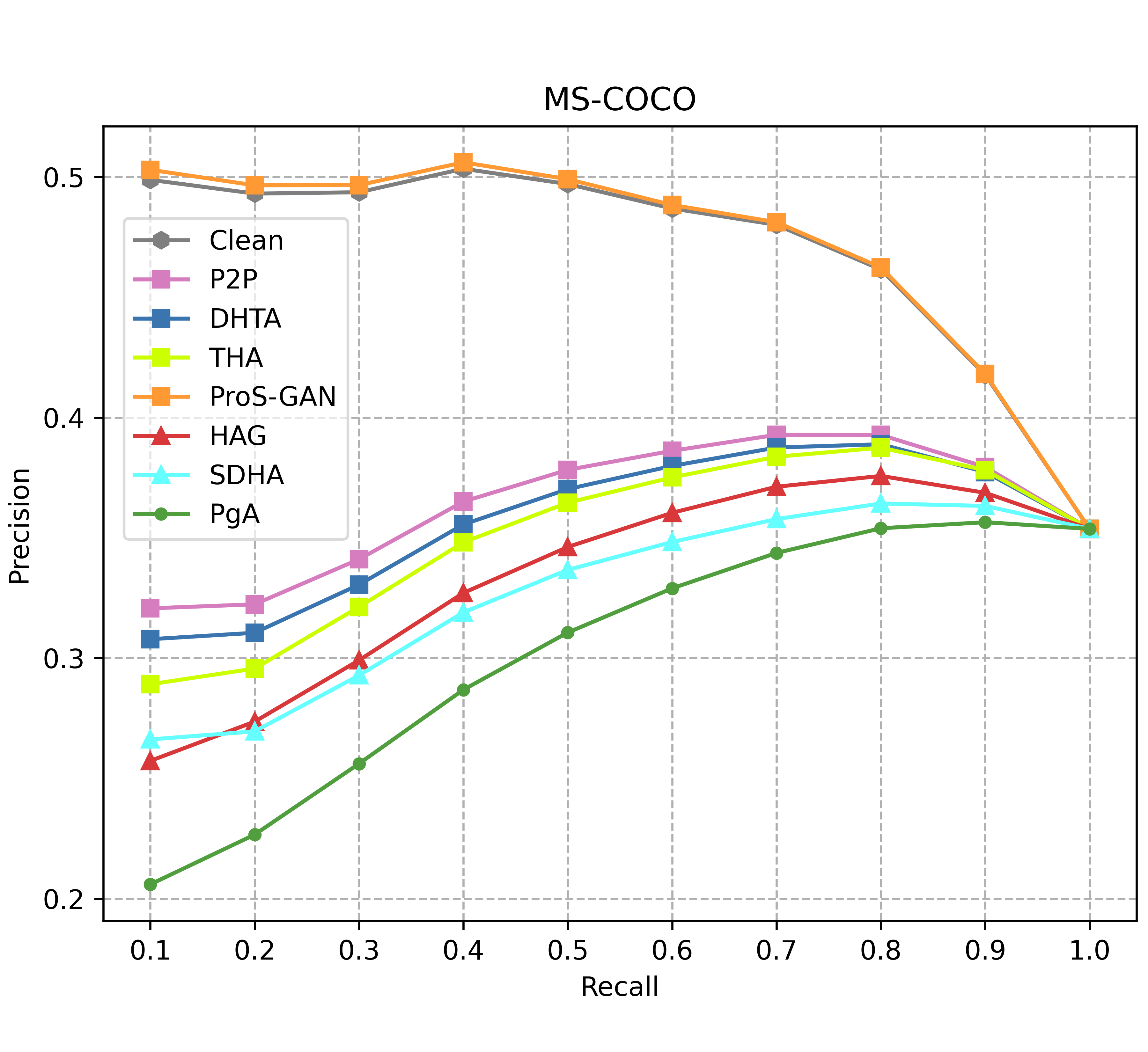}} \\
{\includegraphics[width=0.26\textwidth]{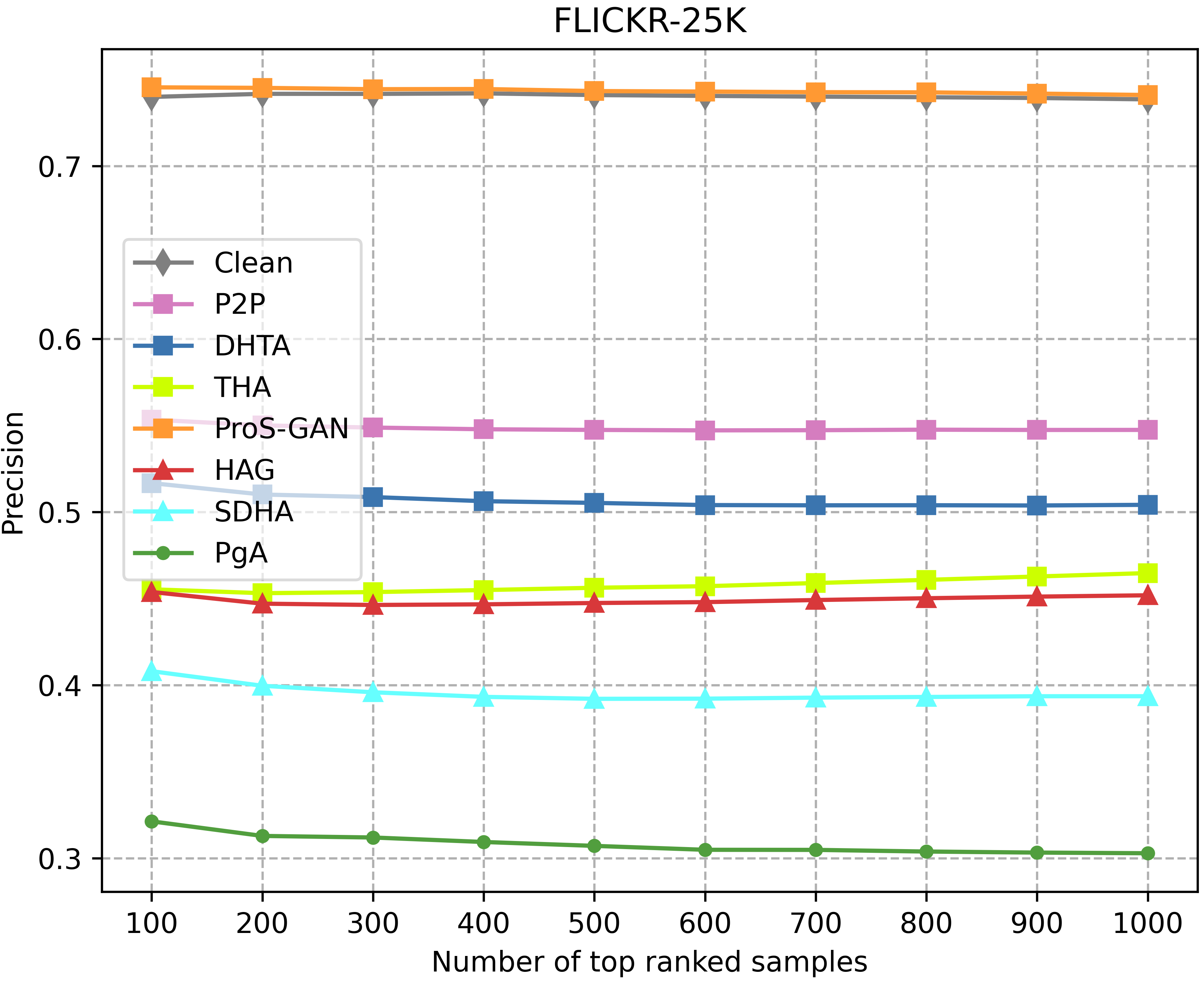}}
\hspace{0.2cm}
{\includegraphics[width=0.26\textwidth]{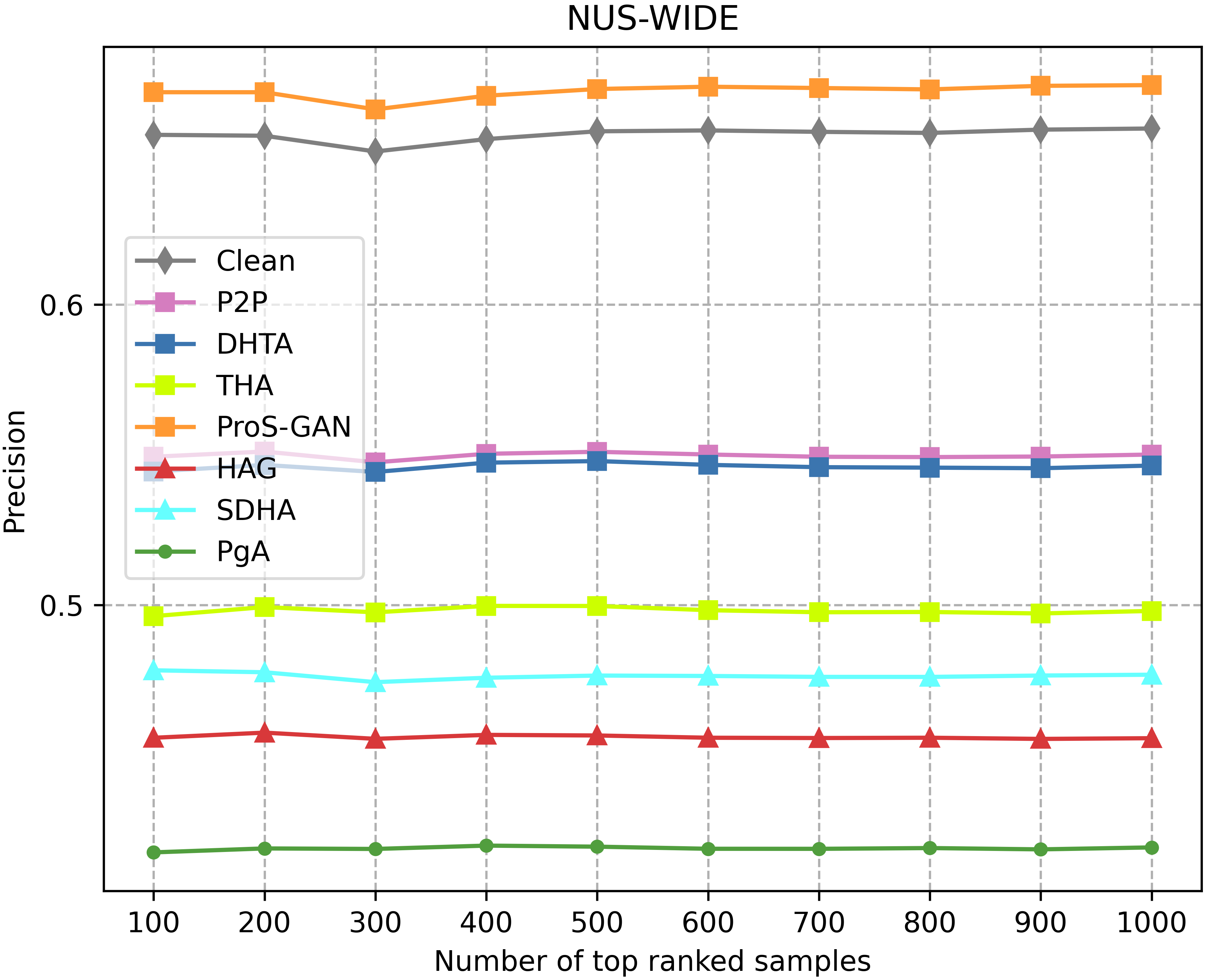}}
\hspace{0.2cm}
{\includegraphics[width=0.26\textwidth]{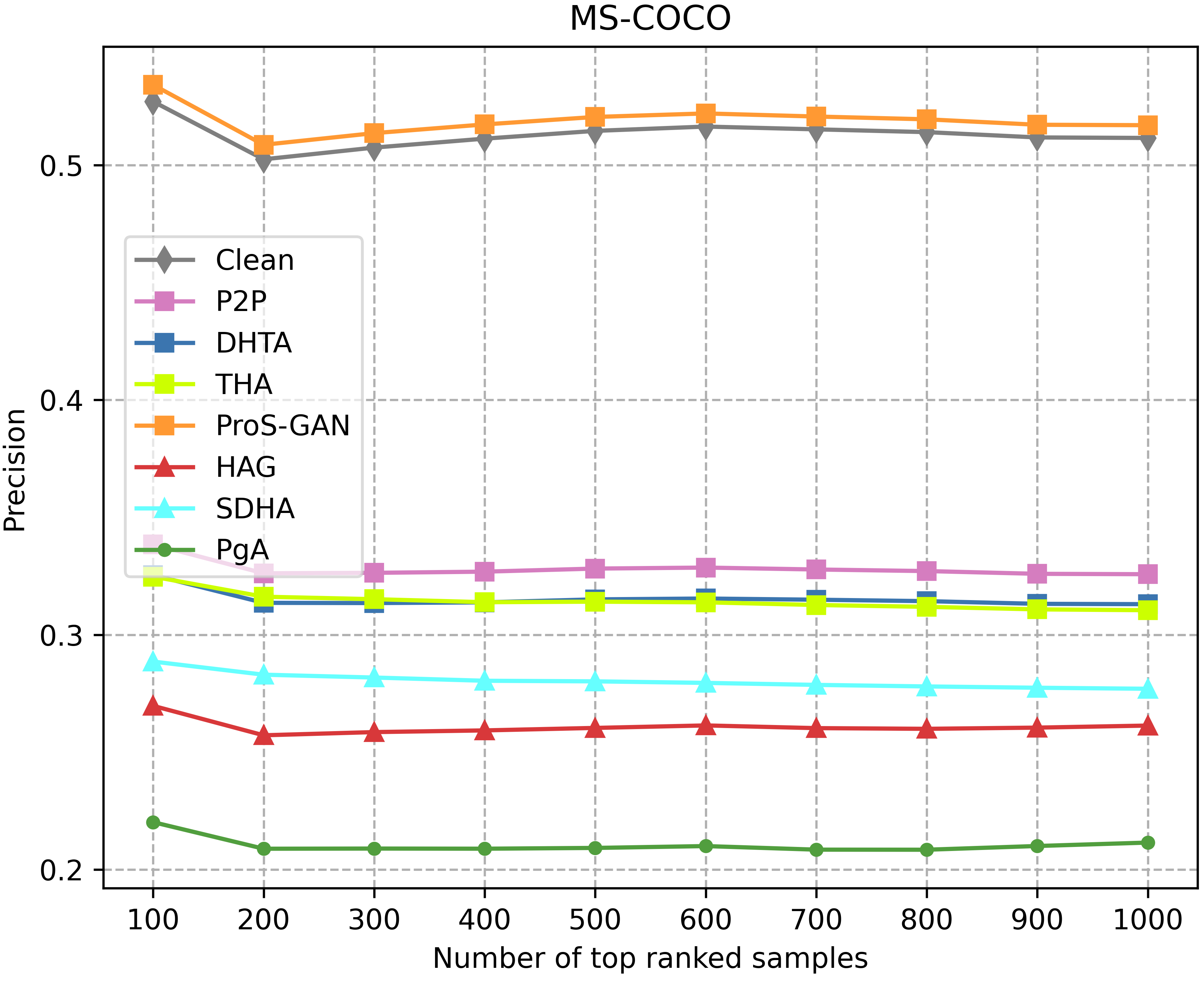}}
\vspace{-.2cm}
\caption{\small Precision-Recall and Precision@topN curves for ATRDH under 32 bits code length.}
\vspace{-0.2cm}
\label{fig:pr_atrdh}
\end{figure*}

\textbf{Protocols.}
To evaluate the performance of PgA, we conduct experiments on the standard metric, \textit{i.e.}, Mean Average Precision (MAP), Precision-Recall (PR) and Precision@topN (P@N) curves \cite{yang2018adversarial}. Following \cite{bai2020targeted}, we calculate MAP values on the top 5,000 results from the database.

\textbf{Baselines.}
Following \cite{yang2018adversarial,bai2020targeted}, we adopt {DPH} as the default attacked hashing method, which is a generic algorithm in deep hashing-based retrieval. AlexNet \cite{krizhevsky2012imagenet} is selected as the default backbone network to implement hashing models on FLICKR-25K, NUS-WIDE, and MS-COCO. 
We also evaluate the attack performance of our method against the defense model trained by ATRDH \cite{wang2021targeted} (the only adversarial training algorithm in deep hashing). We compare the proposed algorithm with multiple hashing attack methods, including P2P, DHTA, ProS-GAN, THA, HAG, and SDHA. For targeted attacks, we randomly select a label as the target label which does not share the same category as the true label. Other details of these methods are consistent with the original literature.


\textbf{Implementation Details.}
We use stochastic gradient descent (SGD) for target hashing models with the initial learning rate of $0.01$ and the momentum $0.9$ as optimizers. We fix the batch size of images as $32$ and the weight decay parameter as $5\times10^{-4}$. All images are resized to $224 \times 224$ and normalized in $[0,1]$ before feeding into hashing models.
For the proposed attack method PgA, we adopt PGD \cite{bai2020targeted} to optimize adversarial examples. The step size $\eta$ and the number of iterations $T$ are set to $1/255$ and $100$, respectively. 
The perturbation budget $\epsilon$ is set to $8/255$. All codes are based on PyTorch 1.12 and are executed on NVIDIA RTX 3090 GPUs.

\begin{table*}[t]
\centering
\begin{minipage}[t]{0.42\textwidth}
\centering
\makeatletter\def\@captype{table}\makeatother
\caption{\small MAP (\%) and time (second per image) of hashing attack methods for hashing models with 32-bit hash code length.}
\label{table:eff}
\vspace{-0.2cm}
\resizebox{\textwidth}{!}{
\begin{tabular}{lcccccc}
\toprule
~ & \multicolumn{2}{c}{FLICKR-25K}& \multicolumn{2}{c}{NUS-WIDE}& \multicolumn{2}{c}{MS-COCO} \\
\cmidrule(r){2-3} \cmidrule(r){4-5} \cmidrule{6-7}
Method &MAP &Time &MAP &Time &MAP &Time \\
\midrule
P2P &42.69 &0.59 &31.76 &0.58 &21.78 &0.58  \\
DHTA &32.63 &0.59 &26.01 &0.58 &19.23 &0.59 \\
ProS-GAN &76.31 &5.10 &34.63 &5.03 &53.67 &2.08  \\
THA &37.54 &0.13 &27.21 &0.08 &20.65 &0.07   \\
HAG &24.37 &0.59 &14.32 &0.58 &13.59 &0.59   \\
SDHA &19.63 &1.68 &15.36 &1.77 &12.36 &1.68  \\
$\text{PgA}^\dagger$ (Ours) &15.49 &\textbf{0.04} &{11.93} &\textbf{0.04} &{10.56} &\textbf{0.04}  \\
PgA (Ours) &\textbf{15.18} &\textbf{0.04} &\textbf{11.53} &\textbf{0.04} &\textbf{9.41} &\textbf{0.04}  \\
\bottomrule
\end{tabular}}
\end{minipage}
\hspace{0.5cm}
\begin{minipage}[t]{0.42\textwidth}
\centering
\makeatletter\def\@captype{table}\makeatother
\caption{\small MAP (\%) and time (second per image) of hashing attack methods for ATRDH trained hashing models with 32 bits.}
\label{table:eff_atrdh}
\vspace{-0.2cm}
\resizebox{\textwidth}{!}{
\begin{tabular}{lcccccc}
\toprule
~ & \multicolumn{2}{c}{FLICKR-25K}& \multicolumn{2}{c}{NUS-WIDE}& \multicolumn{2}{c}{MS-COCO} \\
\cmidrule(r){2-3} \cmidrule(r){4-5} \cmidrule{6-7}
Method &MAP &Time &MAP &Time &MAP &Time \\
\midrule
P2P &54.56 &0.59 &56.70 &0.59 &33.24 &0.58  \\
DHTA &53.26 &0.58 &56.50 &0.59 &31.62 &0.58 \\
ProS-GAN &73.33 &5.09 &67.18 &5.13 &51.30 &2.07  \\
THA &49.94 &0.13 &50.92 &0.08 &30.52 &0.06   \\
HAG &45.58 &0.59 &47.46 &0.59 &27.49 &0.58   \\
SDHA &40.68 &1.60 &47.16 &1.66 &27.75 &1.60  \\
$\text{PgA}^\dagger$ (Ours) &34.67 &\textbf{0.04} &{42.84} &\textbf{0.04} &{21.74} &\textbf{0.04}  \\
PgA (Ours) &\textbf{33.80} &\textbf{0.04} &\textbf{42.22} &\textbf{0.04} &\textbf{21.42} &\textbf{0.04}  \\
\bottomrule
\end{tabular}}
\end{minipage}
\end{table*}

\begin{table*}[t]
\scriptsize
\begin{center}
\caption{\small MAP (\%) / time (second per image) on attacking hashing models with 32-bit hash code length. The tested attack methods all adopt PGD \cite{madry2017towards} to construct adversarial samples with 20 steps (\textit{i.e.}, $T=20$ and $\eta=1/255$).}
\vspace{-2mm}
\label{table:eff_pgd}
\setlength{\tabcolsep}{4.0mm}
\resizebox{1.0\textwidth}{!}{
\begin{tabular}{lcccccc}
\toprule
~ & \multicolumn{2}{c}{FLICKR-25K}& \multicolumn{2}{c}{NUS-WIDE}& \multicolumn{2}{c}{MS-COCO} \\
\cmidrule(r){2-3} \cmidrule(r){4-5} \cmidrule{6-7}
Method &DPH &ATRDH &DPH &ATRDH &DPH &ATRDH \\ 
\midrule
P2P &41.66/0.01 &58.12/0.01 &33.02/0.01 &57.91/0.01 &22.37/0.01 &33.27/0.01  \\
DHTA &34.21/0.01 &55.19/0.01 &26.91/0.01 &58.26/0.01 &19.45/0.01 &32.79/0.01  \\
THA &40.67/0.10 &53.48/0.10 &30.33/0.05 &52.80/0.06 &20.92/0.04 &31.40/0.03   \\
HAG &24.95/0.01 &52.42/0.01 &15.36/0.01 &48.93/0.01 &15.16/0.01 &27.69/0.01   \\
SDHA &24.71/0.03 &43.72/0.03 &18.33/0.02 &49.12/0.02 &15.43/0.02 &27.94/0.02  \\
PgA (Ours) &\textbf{16.17}/0.01 &\textbf{37.58}/0.01 &\textbf{12.44}/0.01 &\textbf{43.81}/0.01 &\textbf{12.46}/0.01 &\textbf{22.47}/0.01  \\ 
\bottomrule
\end{tabular}
}
\end{center}
\vspace{-.3cm}
\end{table*}

\subsection{Attack Results}
Table \ref{table:map_hash} and Table \ref{table:map_atrdh} present the attack performance (MAP) of different attack methods on original deep hashing networks without defense and adversarially trained models, respectively. The lower the MAP value, the stronger the attack performance. The "Clean" in these tables is to query with benign images, so MAP values refer to the original retrieval performance of the hashing model without attack. 
From Table \ref{table:map_hash}, we can observe that the proposed method can greatly reduce the MAP values on three datasets with the hash bits varying from 16 to 64 and outperforms all other attacks. Compared to DHTA \cite{bai2020targeted}, the strongest targeted attack in Table \ref{table:map_hash}, our PgA achieves average boosts of 18.68\%, 14.66\%, and 9.03\% for tested bits on FLICKR-25K, NUS-WIDE, and MS-COCO, respectively. Moreover, our method outperforms it by an average of 4.40\%, 4.09\%, and 2.59\% on three datasets, compared with the state-of-the-art non-targeted attack, SDHA.
As for the defense model trained by ATRDH (the only adversarial training algorithm in deep hashing), Table \ref{table:map_atrdh} shows that all the MAP values of PgA are lower than other attack methods. Even in the face of SDHA, the proposed PgA brings an average improvement of 7.70\%, 5.39\%, and 6.02\% for FLICKR-25K, NUS-WIDE, and MS-COCO, respectively.
The superior performance of our method owes to the superiority of the pharos code, which considers the positive and negative samples simultaneously. In contrast, HAG and SDHA merely use the information from benign and positive samples, respectively. Thus, pharos code-based PgA is better than the previous state-of-the-arts.

For a more comprehensive comparison, PR and P@N curves of different methods on three datasets with 32 bits length are shown in Fig. \ref{fig:pr_hash} and \ref{fig:pr_atrdh}. We can see that the curves of our method are always below all others, demonstrating that our algorithm can attack hashing models more effectively.

In addition, we compare the attack performance of PgA with that of PgA$^\dagger$ to evaluate its effectiveness. The results in Table \ref{table:map_hash} and \ref{table:map_atrdh} shows that PgA with Eq. (\ref{eq:eff_pga}) is better than PgA$^\dagger$ with Eq. (\ref{eq:pga}) in most cases. Hence, the accelerated operation in Eq. (\ref{eq:eff_pga}) can effectively improve the attack effect.

\subsection{Efficiency Analysis}
To confirm the high efficiency of the proposed method, we record the MAP and time of various attack methods, where the time denotes the average interval (second per image) to conduct attacks for the test set. It is noted that this time usually includes the training time of the attack model, \textit{e.g.}, ProS-GAN, and THA. The results are summarized in Table \ref{table:eff} and \ref{table:eff_atrdh}. It is observed that our PgA achieves the strongest attack performance and the shortest attack time for all datasets. Tables \ref{table:eff} and \ref{table:eff_atrdh} are similar in terms of time results, and here we mainly focus on Table \ref{table:eff} for our analysis. Specifically, ProS-GAN has the lowest attack efficiency because it requires a few hours to train a generative network for attack. ProS-GAN takes about 127, 125, and 52 times longer than PgA on FLICKR-25K, NUS-WIDE, and MS-COCO, respectively. Moreover, P2P, DHTA, and HAG have similar attack time and they are much faster than ProS-GAN. Nevertheless, since they require 2000 iterations for gradients, they are still more than 13 times slower than our PgA. In summary, PgA not only outperform all the previous methods in attack performance but also can produce adversarial examples with the fastest speed.

To further verify the high efficiency of our PgA in the same setting, we use PGD-20 to optimize adversarial perturbations for all attack methods, as shown in Table \ref{table:eff_pgd}. PgA has the same speed as P2P, DHTA, and HAG, because they can directly calculate the target code to guide the generation of adversarial samples extremely fast. However, THA costs a lot of time to learn to construct the target code with a fully-connected network, so it is much slower than PgA. Furthermore, SDHA is less efficient than PgA because of its complex objective function \cite{lu2021smart}.

\begin{figure*}[t]
\centering 
\begin{subfigure}{0.32\textwidth}
    \includegraphics[width=0.99\textwidth]{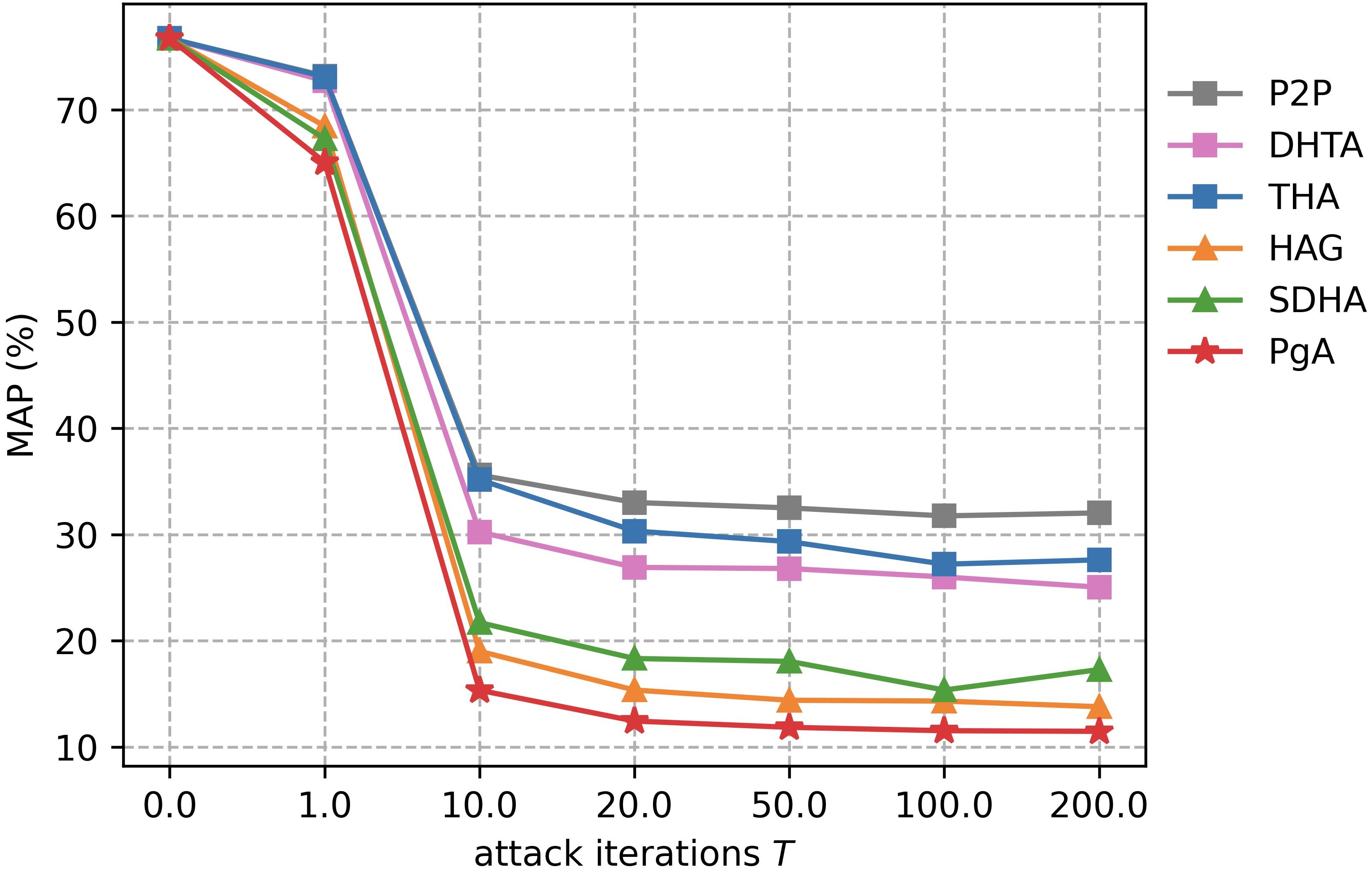}
    \caption{$T$-DPH}
    \label{fig:t_dph}
\end{subfigure}
\begin{subfigure}{0.32\textwidth}
    \includegraphics[width=0.99\textwidth]{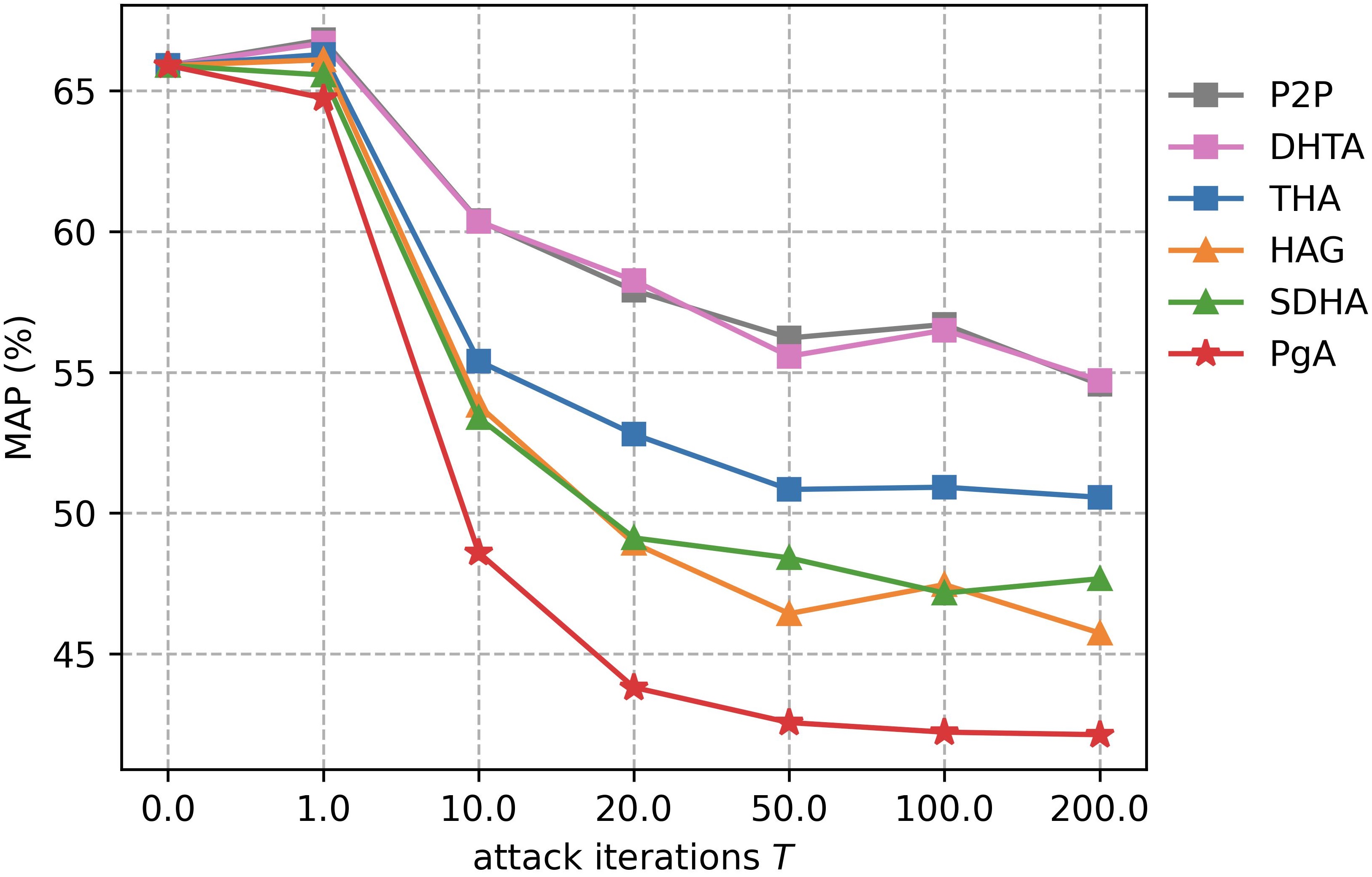}
    \caption{$T$-ATRDH}
    \label{fig:t_atrdh}
\end{subfigure}
\begin{subfigure}{0.32\textwidth}
    \includegraphics[width=1.0\textwidth]{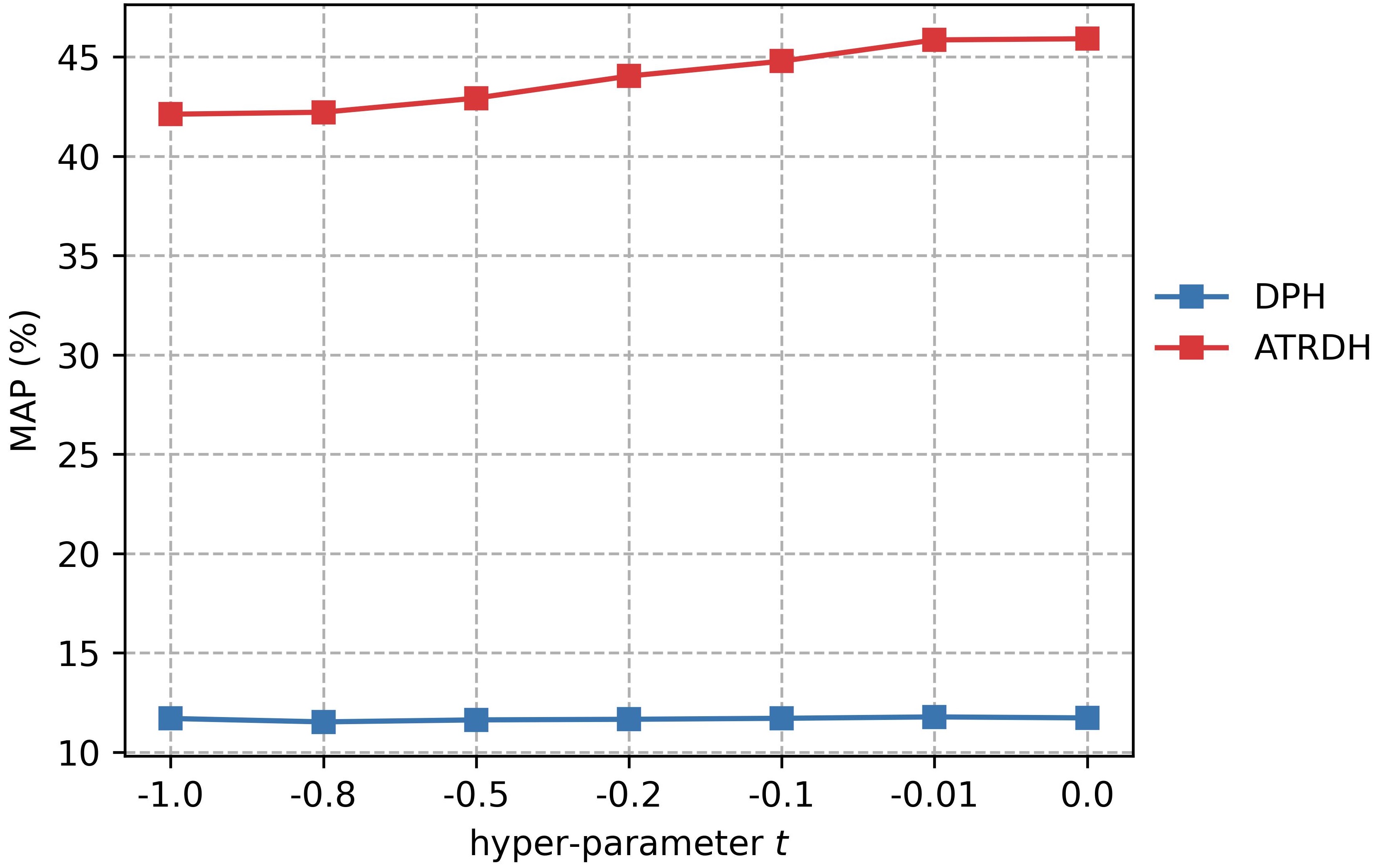}
    \caption{$t$}
    \label{fig:t}
\end{subfigure}
\caption{\small MAP-$T$ on NUS-WIDE for (a) DPH and (b) ATRDH trained model. (c) Attack performance with different $t$ in Eq. (\ref{eq:eff_pga}).}
\label{fig:hyp}
\end{figure*}

\begin{table*}[t]
\scriptsize
\begin{center}
\caption{\small MAP (\%) for different hashing models on NUS-WIDE, including DPSH \cite{li2016feature}, HashNet \cite{cao2017hashnet}, DSDH \cite{li2017deep}, DCH \cite{cao2018deep}, CSQ \cite{yuan2020central} and CSQ-C \cite{doan2022one}, implemented based on VGG11 \cite{simonyan2014very}.}
\vspace{-.3cm}
\label{tab:uni}
\resizebox{1.0\textwidth}{!}{
\begin{tabular}{lcccccccccccc}
\toprule
~ & \multicolumn{6}{c}{No defense}& \multicolumn{6}{c}{ATRDH}  \\
\cmidrule(r){2-7} \cmidrule{8-13}
Method &DPSH &HashNet &DSDH &DCH &CSQ &CSQ-C &DPSH &HashNet &DSDH &DCH &CSQ &CSQ-C \\
\midrule
Clean &82.20 &79.57 &81.90 &77.43 &80.08 &80.81 &70.38 &68.48 &68.77 &70.50 &75.52 &74.97  \\
P2P &29.50 &31.40 &29.63 &37.04 &32.89 &32.64 &48.31 &56.43 &54.35 &41.10 &35.86 &35.89   \\
DHTA &23.38 &25.12 &22.30 &31.68 &27.35 &27.27 &47.96 &54.46 &53.21 &37.96 &31.78 &31.52  \\
ProS-GAN &29.84 &28.62 &30.05 &62.87 &31.01 &33.24 &69.64 &69.33 &70.03 &58.21 &73.72 &72.35  \\
THA &21.79 &19.65 &23.57 &27.18 &27.09 &31.37 &51.14 &57.00 &55.80 &33.58 &34.72 &41.94 \\
HAG &17.18 &14.56 &16.92 &25.28 &19.91 &17.56 &47.17 &48.25 &50.96 &29.47 &18.34 &20.53  \\
SDHA &15.38 &13.31 &16.69 &22.35 &11.78 &13.93 &42.78 &46.75 &47.94 &46.28 &18.77 &20.79  \\
PgA (Ours) &\textbf{10.57} &\textbf{11.94} &\textbf{11.30} &\textbf{15.04} &\textbf{6.86} &\textbf{7.17} &\textbf{37.37} &\textbf{42.31} &\textbf{41.01} &\textbf{21.65} &\textbf{14.83} &\textbf{15.26}  \\
\bottomrule
\end{tabular}
}
\end{center}
\end{table*}

\subsection{Analysis on Hyper-Parameters}
\textbf{Effect of $T$ \& Efficiency.}
Figure \ref{fig:t_dph} and \ref{fig:t_atrdh} present attack performance (MAP) with different attack iterations (\textit{i.e.}, $T$) of PGD. Overall, the MAP values decrease with increasing $T$. When $T$ is greater than 20, the attack performance tends to level off for DPH, and 50 for ATRDH. For the same iterations, the attack performance of PgA maintains a large gap with other methods. Therefore, PgA is an efficient tool for evaluating the robustness of deep hashing networks.

\textbf{Effect of $t$.}
Figure \ref{fig:t} illustrates the effect of hyper-parameter $t$ on the attack performance. For the DPH model without defense, there is no appreciable change in MAP for different $t$. For ATRDH, the attack performance shows a small decrease as $t$ increases. Although attack performance is not extremely sensitive to $t$, picking an appropriate value of $t$ can not be ignored.

\subsection{Universality on Different Hashing Methods}
We argue that the proposed attack algorithm is generic to most popular hashing models. To verify this point, we conduct adversarial attacks on multiple hashing methods with 32-bit hash code length. The results are reported in Table \ref{tab:uni}. It can be seen from the table that our PgA is effective in fooling the illustrated hashing models with better attack performance than others. Firstly, when testing with hashing methods without defense, our PgA exceeds the previous state-of-the-art SDHA in all cases. Especially with DCH, there is a 7.31\% gap between PgA and SDHA. Moreover, under the defense of ATRDH, PgA reduces the MAP of all hashing methods to significant minimums. Also, our PgA brings a 24.63\% enhancement on the DCH model compared to the SDHA. Thus, the above phenomena demonstrate the universality of the proposed attack method, which can be utilized in most popular hashing algorithms. We make another further evaluation on the defense model trained with PgA, and please refer to \cref{ap:adv}.

\section{Conclusion}
In this paper, we proposed the adversarial attack method (\textit{i.e.}, PgA) for efficiently evaluating the adversarial robustness of deep hashing-based retrieval. Specifically, we provided the PGM to fast obtain the pharos code as the optimal representative of the image semantics for the attack in deep hashing. Moreover, PgA took the pharos code as "label" to guide the non-targeted attack, where the similarity between the pharos code and the hash code of adversarial example was minimized. Besides, we added adaptive weighting into the Hamming distance calculation, which further boosts the strength of PgA. Experiments showed that our algorithm performed state-of-the-art attack performance and efficiency compared to the previous attack methods in deep hashing-based retrieval.

{\small
\bibliographystyle{ieee_fullname}
\bibliography{egbib}
}

\clearpage
\appendix

\section{Proof of PGM}
\label{a1}
\textbf{Theorem}
pharos code $\boldsymbol{b}^\star$ which satisfies Eq. (6) can be calculated by the Pharos Generation Method (PGM), \textit{i.e.}, 
\begin{equation*}
    \begin{aligned}
    \boldsymbol{b}^\star &= \arg\min_{\boldsymbol{b}^\star\in\{-1,+1\}^K} \sum_{i}\sum_{j}[w_i {D}_{\rm{H}}(\boldsymbol{b}^\star, \boldsymbol{b}_i^{(\rm{p})}) \\ &- w_j {D}_{\rm{H}}(\boldsymbol{b}^\star, \boldsymbol{b}_j^{(\rm{n})})] \\
    &=\operatorname{sign}\left(\sum_{i}^{N_{\rm{p}}}\sum_{j}^{N_{\rm{n}}}(w_{i}\boldsymbol{b}_i^{(\rm{p})} - w_{j}\boldsymbol{b}_j^{(\rm{n})})\right).
    \end{aligned}
\end{equation*}
\textit{proof.}
We define the following function:
\begin{equation*}
    \begin{aligned}
    \psi(\boldsymbol{b})=\sum_{i}\sum_{j} [w_i {D}_{\rm{H}}(\boldsymbol{b}, \boldsymbol{b}_i^{(\rm{p})}) - w_j {D}_{\rm{H}}(\boldsymbol{b}, \boldsymbol{b}_j^{(\rm{n})})]
    \end{aligned}.
\end{equation*}
As the pharos code $\boldsymbol{b}^\star$ need to be the optimal solution of the minimizing objective, the above theorem is equivalent to prove the following inequality:
\begin{equation*}
    \begin{aligned}
    \psi(\boldsymbol{b})\geq \psi(\boldsymbol{b}^\star),
    \quad \forall~\boldsymbol{b}\in\{-1,+1\}^K
    \end{aligned}.
\end{equation*}
Let $\boldsymbol{b}=\{b_1,b_2,...,b_K\}$, then we have
\begin{equation*}
    \begin{aligned}
    &\psi(\boldsymbol{b})
    =\sum_{i}\sum_{j} [w_i \frac{1}{2}(K-\boldsymbol{b}^\top \boldsymbol{b}_i^{(\rm{p})}) - w_j \frac{1}{2}(K-\boldsymbol{b}^\top \boldsymbol{b}_j^{(\rm{n})})] \\
    =&-\frac{1}{2}\sum_{i}\sum_{j} [w_i\boldsymbol{b}^\top \boldsymbol{b}_i^{(\rm{p})} - w_j\boldsymbol{b}^\top \boldsymbol{b}_j^{(\rm{n})}] + \xi \\
    =&-\frac{1}{2}\sum_{i}\sum_{j} [w_i\sum_{k=1}^K{b}_k{b}_{ik}^{(\rm{p})}-w_j\sum_{k=1}^K{b}_k{b}_{jk}^{(\rm{n})}]+\xi \\
    =&-\frac{1}{2}\sum_{i}\sum_{j} \left(\sum_{k=1}^K w_i{b}_k{b}_{ik}^{(\rm{p})}-\sum_{k=1}^K w_j{b}_k{b}_{jk}^{(\rm{n})}\right) + \xi \\
    =&-\frac{1}{2}\sum_{i}\sum_{j} \sum_{k=1}^K{b}_k( w_i{b}_{ik}^{(\rm{p})}-w_j{b}_{jk}^{(\rm{n})}) + \xi \\
    =&-\frac{1}{2}\sum_{k=1}^K b_k\sum_{i}\sum_{j}(w_ib_{ik}^{(\rm{p})}-w_jb_{jk}^{(\rm{n})})+\xi, \\
    \end{aligned}
\end{equation*}
where $\xi$ is a constant.
Similarly, 
\begin{equation*}
    \begin{aligned}
    \psi(\boldsymbol{b}^\star)=-\frac{1}{2}\sum_{k=1}^K b^\star_{k}\sum_{i}\sum_{j}(w_ib_{ik}^{(\rm{p})}-w_jb_{jk}^{(\rm{n})})+\xi.
    \end{aligned}
\end{equation*}
Due to the nature of absolute value, we have
\begin{equation*}
    \begin{aligned}
    &\psi(\boldsymbol{b})
    =-\frac{1}{2}\sum_{k=1}^K b_k\sum_{i}\sum_{j}(w_ib_{ik}^{(\rm{p})}-w_jb_{jk}^{(\rm{n})})+\xi \\
    \geq&-\frac{1}{2}\sum_{k=1}^K \left|b_k\sum_{i}\sum_{j}(w_ib_{ik}^{(\rm{p})}-w_jb_{jk}^{(\rm{n})})\right|+\xi \\
    =&-\frac{1}{2}\sum_{k=1}^K\vert{b_k}\vert \left|\sum_{i}\sum_{j}(w_ib_{ik}^{(\rm{p})}-w_jb_{jk}^{(\rm{n})})\right|+\xi \\
    =&-\frac{1}{2}\sum_{k=1}^K \left|\sum_{i}\sum_{j}(w_ib_{ik}^{(\rm{p})}-w_jb_{jk}^{(\rm{n})})\right|+\xi \\
    =&-\frac{1}{2}\sum_{k=1}^K \operatorname{sign}(\sum_{i}\sum_{j}(w_ib_{ik}^{(\rm{p})}-w_jb_{jk}^{(\rm{n})}))\sum_{i}\sum_{j}(w_ib_{ik}^{(\rm{p})}-w_jb_{jk}^{(\rm{n})})+\xi \\
    =&-\frac{1}{2}\sum_{k=1}^K b^\star_{k}\sum_{i}\sum_{j}(w_ib_{ik}^{(\rm{p})}-w_jb_{jk}^{(\rm{n})})+\xi \\
    =&\psi(\boldsymbol{b}^\star).
    \end{aligned}
\end{equation*}
That is, $\psi(\boldsymbol{b})\geq\psi(\boldsymbol{b}^\star)$. Hence, the Theorem is proved.

\section{Attack results on CIFAR-10}
\label{ap:cifar}
Table \ref{tab:cifar} shows the results of the hashing attack methods on the single-label dataset CIFAR-10 \cite{cao2017hashnet}. We can observe that our PgA is a little bit better than the state-of-the-art SDHA for DPH. However, the proposed PgA outperforms HAG and SDHA over 2.23\%. Especially under the case of 64 bits, PgA brings an boost of 4.05\% and 10.19\% compared to HAG and SDHA, respectively. 

\begin{table}[ht]
\scriptsize
\begin{center}
\caption{\small MAP (\%) of attack methods on CIFAR-10.}
\label{tab:cifar}
\resizebox{0.99\columnwidth}{!}{
\begin{tabular}{lcccccc}
\toprule
~ & \multicolumn{3}{c}{DPH}& \multicolumn{3}{c}{ATRDH} \\
\cmidrule(r){2-4} \cmidrule(r){5-7}
Method &16 bits &32 bits &64 bits &16 bits &32 bits &64 bits \\
\midrule
Clean &67.72 &78.00 &79.64 &60.98 &62.74 &63.08 \\
P2P &4.11 &3.44 &2.96 &31.55 &31.94 &32.19   \\
DHTA &2.08 &1.24 &0.91 &29.87 &31.12 &31.04  \\
ProS-GAN &2.93 &6.13 &5.17 &64.14 &66.27 &66.98 \\
THA &2.64 &6.77 &8.42 &31.95 &32.79 &35.06   \\
HAG &0.95 &1.16 &1.60 &16.41 &16.51 &18.30   \\
SDHA &{0.32} &0.52 &0.55 &18.90 &20.75 &24.54  \\
PgA (Ours) &\textbf{0.31} &\textbf{0.49} &\textbf{0.48} &\textbf{14.18} &\textbf{13.48} &\textbf{14.25}  \\
\bottomrule
\end{tabular}
}
\end{center}
\end{table}

\section{Adversarial Training}
\label{ap:adv}
We use the generated adversarial samples for adversarial training to verify whether the proposed method is still valid. The objective of the adversarial training is formulated as follows:
\begin{equation}
    \begin{aligned}
        \min_{\theta}\mathcal{L}_{adv}=\mathcal{L}_{ori} -\sum_{i=1 }^N\frac{1}{K}(\boldsymbol{b}^\star_{i})^{\top}f_{\theta}(\boldsymbol{x}_i^\prime)
    \end{aligned},
    \label{eq:obj_adv}
\end{equation}
where $\boldsymbol{b}^\star_{i}$ is the pharos code of the instance $\boldsymbol{x}_i$, and $\boldsymbol{x}_i^\prime$ is the adversarial example of $\boldsymbol{x}_i$. The latter term in Eq. (\ref{eq:obj_adv}) can rebuild similarity between the adversarial sample and the true semantics. $\mathcal{L}_{ori}$ is the original loss function of the deep hashing model, which ensures the basic performance of hashing learning. The experimental results are illustrated in Table \ref{table:adv_train}. The adversarial training does improve the defense capability of the deep hashing model, but our attack method is still valid and significantly outperforms the other methods.

\begin{table}[h]
\scriptsize
\begin{center}
\caption{\small MAP (\%) of attack methods on NUS-WIDE.}
\label{table:adv_train}
\resizebox{0.45\textwidth}{!}{
\begin{tabular}{lccc}
\toprule
Method &16 bits &32 bits &64 bits \\
\midrule
Clean &70.51 &68.50 &62.34 \\
P2P &45.50 &53.08 &56.78   \\
DHTA &43.12 &50.30 &55.47       \\
ProS-GAN &64.27 &67.81 &62.49   \\
THA &48.36 &55.74 &59.90        \\
HAG &45.26 &51.32 &52.26        \\
SDHA &34.67 &45.28 &51.03       \\
PgA (Ours) &\textbf{26.72} &\textbf{37.70} &\textbf{49.70}   \\
\bottomrule
\end{tabular}
}
\end{center}
\end{table}

\end{document}